\UseRawInputEncoding
\documentclass[lettersize,journal]{IEEEtran}
\usepackage{cite}
\usepackage{amsmath}
\usepackage[ruled]{algorithm2e}
\usepackage{float}
\usepackage{url}
\usepackage{xcolor}
\usepackage{graphicx}
\usepackage{enumitem}
\usepackage{multirow}
\usepackage{setspace}
\usepackage[utf8]{inputenc}
\usepackage{hyperref} 
\usepackage[T1]{fontenc} 
\usepackage{booktabs}
\usepackage{bbding}
\usepackage{diagbox}
\usepackage[caption=false,font=footnotesize,labelfont=rm,textfont=rm]{subfig}
\begin{document}
\graphicspath{{}}

\title{Viewport Prediction for Volumetric Video Streaming by Exploring Video Saliency and User Trajectory Information

}

\author{Jie Li,~\IEEEmembership{Member,~IEEE,}
        Zhixin Li,
        Zhi Liu,~\IEEEmembership{Senior Member,~IEEE},
        Peng Yuan Zhou,~\IEEEmembership{Member,~IEEE},
        Richang Hong,~\IEEEmembership{Member,~IEEE},
        Qiyue Li,~\IEEEmembership{Senior Member,~IEEE},
        Han Hu,~\IEEEmembership{Member,~IEEE}
        	
\thanks{Jie Li, Zhixin Li and Richang Hong are with the School of Computer Science and Information Engineering, Hefei University of Technology, Hefei, China (e-mail: lijie@hfut.edu.cn; lizhixin@mail.hfut.edu.cn; hongrc.hfut@gmail.com).}
\thanks{Zhi Liu is with The University of Electro-Communications, Japan (e-mail: liuzhi@uec.ac.jp).}
\thanks{Peng Yuan Zhou is with Aarhus University, Denmark. (e-mail: pengyuan.zhou@ece.au.dk).}
\thanks{Qiyue Li is with the School of Electrical Engineering and Automation, Hefei University of Technology, Hefei, China, and also with the Engineering Technology Research Center of Industrial Automation of Anhui Province, Hefei, China (e-mail: liqiyue@mail.ustc.edu.cn)}
\thanks{Han Hu is with the School of Information and Electronics, Beijing Institute of Technology, Beijing, China (e-mail: huhan627@gmail.com).}
\thanks{Corresponding author: Zhi Liu.}
}
\maketitle

\begin{abstract}

Volumetric video, also referred to as hologram video, is an emerging medium that represents 3D content in extended reality. As a next-generation video technology, it is poised to become a key application in 5G and future wireless communication networks. Because each user generally views only a specific portion of the volumetric video, known as the viewport, accurate prediction of the viewport is crucial for ensuring an optimal streaming performance. Despite its significance, research in this area is still in the early stages.
To this end, this paper introduces a novel approach called Saliency and Trajectory-based Viewport Prediction (STVP), which enhances the accuracy of viewport prediction in volumetric video streaming by effectively leveraging both video saliency and viewport trajectory information.
In particular, we first introduce a novel sampling method, Uniform Random Sampling (URS), which efficiently preserves video features while minimizing computational complexity. 

Next, we propose a saliency detection technique that integrates both spatial and temporal information to identify visually static and dynamic geometric and luminance-salient regions.
Finally, we fuse saliency and trajectory information to achieve more accurate viewport prediction.
Extensive experimental results validate the superiority of our method over existing state-of-the-art schemes. To the best of our knowledge, this is the first comprehensive study of viewport prediction in volumetric video streaming. We also make the source code of this work publicly available.

\end{abstract}


\section{Introduction}
\label{sec:intro}

Volumetric video, also known as holographic video, allows users to fully immerse themselves in a 3D scene and move freely in any direction, providing a 6 degrees of freedom (6DoF) experience \cite{mieloch2022overview,mekuria2016design}. 
Typically, users watch only a section of a video, referred to as a ``viewport''\footnote{Viewport, also referred to as the field of view (FoV), is the portion of a video that a user actually watches. In this paper, we use ``viewport'' and ``FoV'' interchangeably.}, at a time and are allowed to switch between viewports freely at will.
This versatility suggests that volumetric video is expected to become the next-generation video technology for entertainment, education, and manufacturing applications, among others \cite{9537928,han2020vivo}.

\begin{figure*}[!htb]
     \centering
      \includegraphics[width=0.6\linewidth]{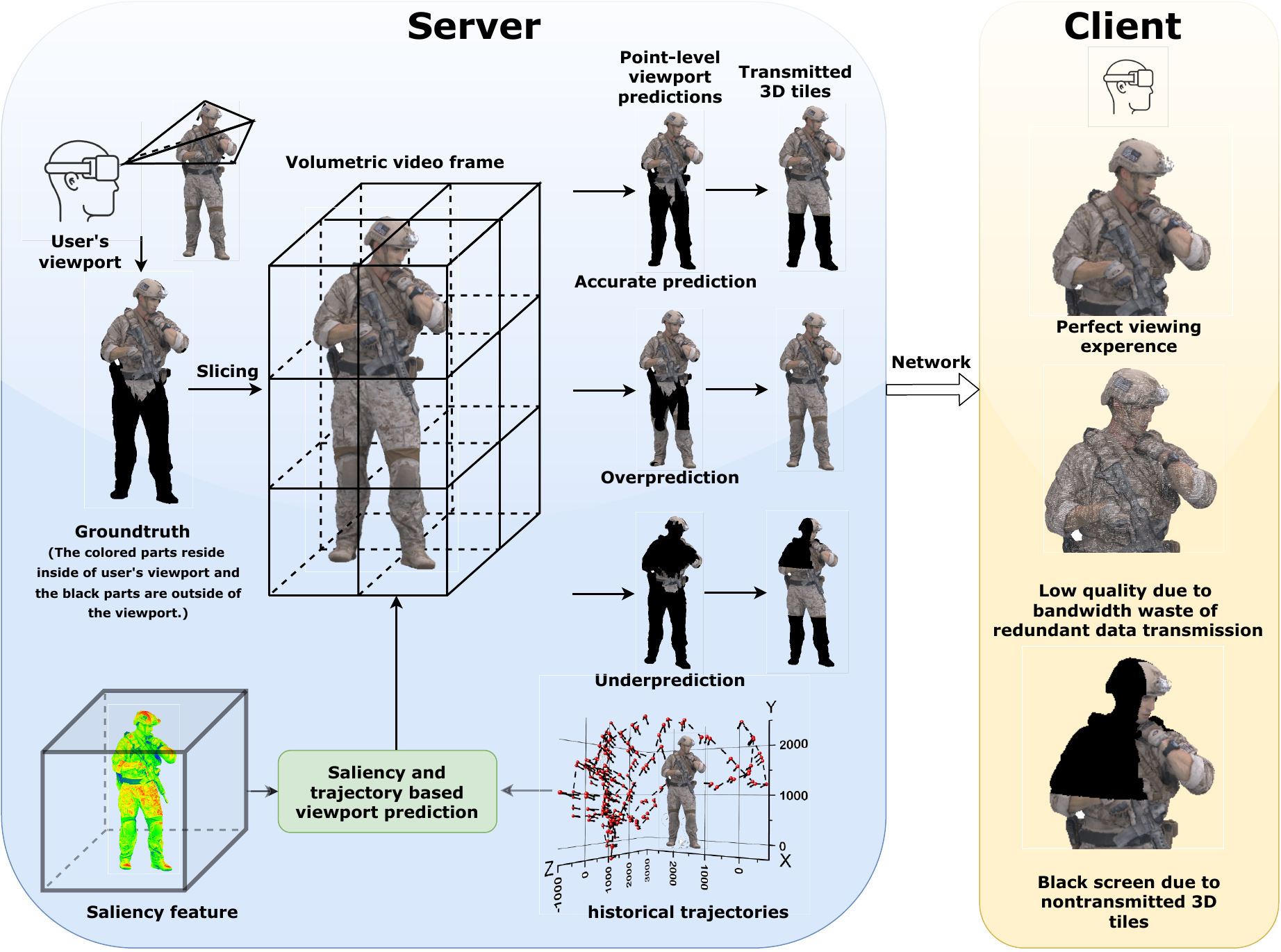}
     \caption{Illustration of tiling and FoV-based adaptive point cloud video streaming. Accurate FoV prediction can lead to a perfect viewing experience, bad prediction results in wasted bandwidth or freeze.}
	  \label{fig_illustration_viewport_prediction}
\end{figure*}

Point cloud video is a prevalent kind of volumetric video, presenting various challenges due to the high bandwidth and low latency requirements for transmission \cite{chen2024dynamic,chen2024progressive,zhou2024blind}.
Many works adopt tiling and adaptive streaming technologies to address these limitations \cite{10.1145/1943552.1943572} by allowing the server to transmit only the video blocks\footnote{Note that volumetric video, or point cloud video, must be partitioned into small blocks for streaming; these blocks are referred to as tiles in some studies. In this work, we use the terms blocks and tiles interchangeably.} within the user's viewport \cite{li2022optimal1}.
However, an inaccurate estimate of the viewport may may result in a degraded viewing experience. Hence, accurate viewport prediction is critical for optimal point cloud video transmission, as illustrated in Fig. \ref{fig_illustration_viewport_prediction}.

When watching point cloud videos, users are typically drawn to visually salient content, such as vibrant colors, noticeable geometric shapes, and motion. Relatedly, the user trajectory consists of the head position and orientation and exhibits temporal coherence, which can be predicted using deep learning techniques\cite{9234536}.
Therefore, the saliency features of a point cloud video and the user trajectory can be used to accurately predict the user's viewport.
Depending on the information used for prediction, there are usually three alternatives for performing viewport prediction for a point cloud video.

\textbf{(1) Trajectory-based viewport prediction.} 
The user's trajectory exhibits temporal coherence, allowing prediction of the next state based on the preceding historical trajectory.

\textbf{(2) Saliency-based viewport prediction.} Visual saliency detection is designed to identify areas that users are likely to view \cite{9097197}. Therefore, visual saliency detection is used for predicting user viewports. The investigation of saliency prediction in 360-degree videos has seen considerable advancement \cite{8578612,9606880,8374803,9729212}. However, to the best of our knowledge, there is still a gap in dynamic saliency detection for 6DoF volumetric video systems.

\textbf{(3) Saliency and Trajectory based hybrid prediction.} To further enhance prediction accuracy, multimodal methodscombining video saliency and trajectory have been proposed \cite{10.1145/3240508.3240669,9416230,7840720}.

Despite significant efforts to predicting user viewports, viewport prediction in volumetric video systems is still in its early stages. The technical challenges are as follows:

\textbf{Lack of efficient techniques in the temporal feature extraction for large-scale point cloud videos:}
The dense nature of point cloud videos results in high computational demands. For the temporal feature extraction in large-scale point cloud videos, it is more efficient to calculate the motion intensity of moving parts rather than process all points. This helps reduce computational costs. However, current methods still face challenges in identifying efficient techniques that prioritize key temporal features, particularly in dense point cloud videos. Many approaches either fail to prioritize critical temporal information or overlook the need for computational efficiency \cite{10122134}. Therefore, it is crucial to design an efficient temporal feature extraction method.

Deficiencies in sampling methods for preserving temporal information in large-scale point cloud videos:
When training a network for saliency detection in point cloud videos, it is crucial to identify a sampling method specifically tailored to point cloud videos. The goal is to reduce the number of points while preserving both temporal and spatial information, while maintaining lower time complexity. Currently, existing heuristic sampling methods \cite{10.5555/3295222.3295263,groh2018flex,8578570,10025770}, primarily designed for static point cloud images, may not be directly applicable to point cloud videos. 
Meanwhile, some heuristic sampling methods sacrifice temporal information, while others suffer from computational complexity (explained in Section \ref{sec:related work}).
Hence, it is important to design an appropriate sampling method to improve the point cloud video sampling efficiency.

\textbf{Inadequate encoding techniques for capturing local visual saliency discrepancies:}
Additionally, in a point cloud video frame, regions with significant geometric variations and brightness changes are particularly captivating to viewers. The current spatial aggregator \cite{8579077,8954075,8954040,10025770,9189936} focuses on encoding local point coordinates to extract geometric features, but neglects the importance of local luminance information. Wang et al. \cite{10301577} focused on static re-identification, which fails to capture dynamic changes. This limitation prevents an accurate representation of spatial saliency features within a single frame of a point cloud video. Therefore, it is essential to design a spatial encoder that integrates both geometric and luminance information from a single video frame.

This paper proposes a high-precision viewpoint prediction network based on neural networks such as LSTM to address these challenges.
Initially, a low-complexity 3D-block based uniform random sampling (URS) unit is introduced. This unit combines spatial uniform partitioning, random sampling, and the K-nearest neighbor (KNN) method to preserve as much temporal and spatial information as possible during the sampling process of the point cloud video. By improving the efficiency of our sampling process, the results for adjacent frames help confirm the feasibility of the saliency detection module.
\begin{table*}[]
\centering
\caption{RELATED VIEWPORT PREDICTION WORK COMPARISON}
\scalebox{0.7}{
\begin{tabular}{clcccc}
\specialrule{0em}{3pt}{3pt}
\hline
\specialrule{0em}{3pt}{3pt}
\textbf{Reference} &
  \multicolumn{1}{c}{\textbf{Video Type}} & \textbf{Degree of Freedom} & 
  \textbf{Using head trajectory} &
  \textbf{Using video saliency} &
  \textbf{Network}  \\\specialrule{0em}{3pt}{3pt} \hline \specialrule{0em}{1pt}{1pt}
\cite{DBLP:conf/mobicom/QianJHG16} &
  \multicolumn{1}{c}{360${^\circ}$ video} & 3DoF &
  \CheckmarkBold &
  \XSolidBold &
  WLR \\ \specialrule{0em}{1pt}{1pt} \hline \specialrule{0em}{1pt}{1pt}
\cite{8486606}                 & \multicolumn{1}{c}{360${^\circ}$ video} & 3DoF           & \CheckmarkBold & \XSolidBold    & LR\&KNN     \\ \specialrule{0em}{1pt}{1pt} \hline \specialrule{0em}{1pt}{1pt}
\cite{10.1145/3083165.3083180} & \multicolumn{1}{c}{360${^\circ}$ video}  &3DoF        & \CheckmarkBold & \CheckmarkBold & LSTM        \\\specialrule{0em}{1pt}{1pt} \hline \specialrule{0em}{1pt}{1pt}
\cite{10.1145/3240508.3240669} &
  \multicolumn{1}{c}{360${^\circ}$ video} & 3DoF &
\CheckmarkBold &
  \CheckmarkBold &
  PanoSalNet \\ \specialrule{0em}{1pt}{1pt} \hline \specialrule{0em}{1pt}{1pt}
\cite{9416230} &
  \multicolumn{1}{c}{360${^\circ}$ video} & 3DoF &
  \CheckmarkBold &
  \CheckmarkBold & Graph Learning \\\specialrule{0em}{1pt}{1pt} \hline \specialrule{0em}{1pt}{1pt}
\cite{9234536}                 & \multicolumn{1}{c}{VR} &6DoF & \CheckmarkBold & \XSolidBold    & LSTM\&MLP \\\specialrule{0em}{1pt}{1pt} \hline \specialrule{0em}{1pt}{1pt}
\specialrule{0em}{1pt}{1pt}
\textbf{Our}                & \multicolumn{1}{c}{\textbf{Point cloud video}} &\textbf{6DoF} & \CheckmarkBold & \CheckmarkBold    & \textbf{LSTM\&Saliency Detection} \\\specialrule{0em}{1pt}{1pt} \hline 
\end{tabular}}
\label{tb:ref}
\end{table*}
This paper also proposes a method for extracting temporal saliency features through local feature comparison. This is achieved using a temporal contrast (TC) layer, which reduces computational efforts and eliminates network instability caused by the unordered nature of the point cloud.

We introduce an encoder called the local discrepancy catcher (LDC), to extract spatial salient features from each frame of the point cloud video. 

We term the approach that integrates spatial-temporal saliency detection and LSTM-based user trajectory estimation for multi-modal viewport prediction as  Saliency and Trajectory-based Viewport
Prediction (STVP). This fusion of trajectory and saliency accounts for the influence of video content and the user's trajectory on the dynamic viewport, improving viewport prediction estimates.
In summary, the main contributions of this paper include:

\begin{itemize}
\item the formulation of an efficient sampling method (URS) that reduces the computational load while preserving essential video features.
\item the enhancement of the current saliency detection method by integrating temporal and spatial information, facilitating the capture of visual salient regions.
\item the implementation of a deep fusion strategy to integrate saliency and trajectory information, leading to more precise viewport prediction.
\item the extensive experiments validating the superiority of our method over existing state-of-the-art schemes.
\end{itemize}

The paper is structured as follows: Section \ref{sec:overview} outlines the user viewport prediction model. Section \ref{sec:saliency} discusses the principles of saliency models for point cloud videos. Section \ref{sec:LSTM} explains the principles of LSTM-based user head trajectory prediction and the feature fusion mechanism. Section \ref{sec:experimental} covers the experimental setup and performance evaluation results. Finally, Section \ref{sec:summary} concludes the paper.

\section{Related WORK}\label{sec:related work}

This section describes related work on saliency detection, viewport prediction and point cloud sampling.

\subsection{Viewport Prediction}

Efficiently predicting the viewer's FoV is crucial for immersive video experiences, such as 360-degree and point cloud videos. Linear regression (LR) \cite{10.1145/3123266.3123291,DBLP:conf/mobicom/QianJHG16} is commonly used, along with variations like weighted LR, KNN-based methods, and the incorporation of confidence values \cite{DBLP:conf/mobicom/QianJHG16,8486606,8642510} to enhance accuracy. For example, S. Gül et al. \cite{gul2020low} employ LR to predict each of the 6DoF, resulting in two-dimensional video outputs. Y. S. de la Fuente et al. \cite{8642510} predict future positions by monitoring the angular velocity and acceleration of head movements.

With the advancement of neural networks, these techniques have been applied to viewport prediction. For instance, S. Yoon et al. propose a 6DoF AR algorithm for visual consistency, while X. Hou et al. employ LSTM and MLP models for low-latency 6DoF VR with tracking devices \cite{9730327,9234536}.

However, these methods only consider the temporal characteristics of user head motion and overlook the impact of video content on user attention, which can degrade prediction performance.

By incorporating video content features, viewport prediction can be significantly improved.  For instance, C. Fan et al. \cite{10.1145/3083165.3083180} combine 360-degree video content features with head-mounted display (HMD) sensor data to predict user fixation. Similarly, studies \cite{10.1145/3240508.3240669, 9606880, li2022spherical} leverage  360-degree video saliency and head orientation history to predict 360-degree video viewport.

However, as shown in Table \ref{tb:ref}, research on viewport prediction for point cloud videos, which considers both video content and user trajectory, is limited compared to 360-degree or VR videos.
In this study, we aim to bridge this gap by jointly considering point cloud video saliency and user head trajectory information, thereby maximizing viewport prediction performance.

\subsection{Visual Saliency Detection}
Visual saliency is a prominent research topic in computer vision, aiming to quickly identify the most {attention-grabbing objects or patterns in images or videos. This is crucial for predicting the user's visual attention and viewport \cite{730558,10.1145/3240508.3240669,5459462,li2022spherical,li2022optimal,8466906}.

Visual saliency detection can be divided into two categories: imag-based and video-based detection. Deep learning techniques enhance saliency detection in still images, overcoming related challenges with various methods. 
For instance, Q. Lai et al. \cite{9749931} propose a weakly supervised method for visual saliency prediction using a joint loss module and a deep neural network. M. Zhong et al. \cite{9853271} develop a computational visual model based on support vector regression to predict user attention to specific categories. G. Ma et al. \cite{9395398} revisit image salient object detection from a semantic perspective, providing a novel network to learn the relative semantic saliency of two input object proposals.
In the domain of 3D point clouds, V. F. Figueiredo et al. \cite{9287102} use orthographic projections and established saliency detection algorithms to generate a 3D point cloud saliency map. In contrast, X. Ding et al. \cite{8726371} propose a novel approach for detecting saliency in point clouds by combining local distinctness and global rarity cues, supported by psychological evidence.

\begin{figure*}[htb!]
    \centering
    \includegraphics[width=0.8\linewidth]{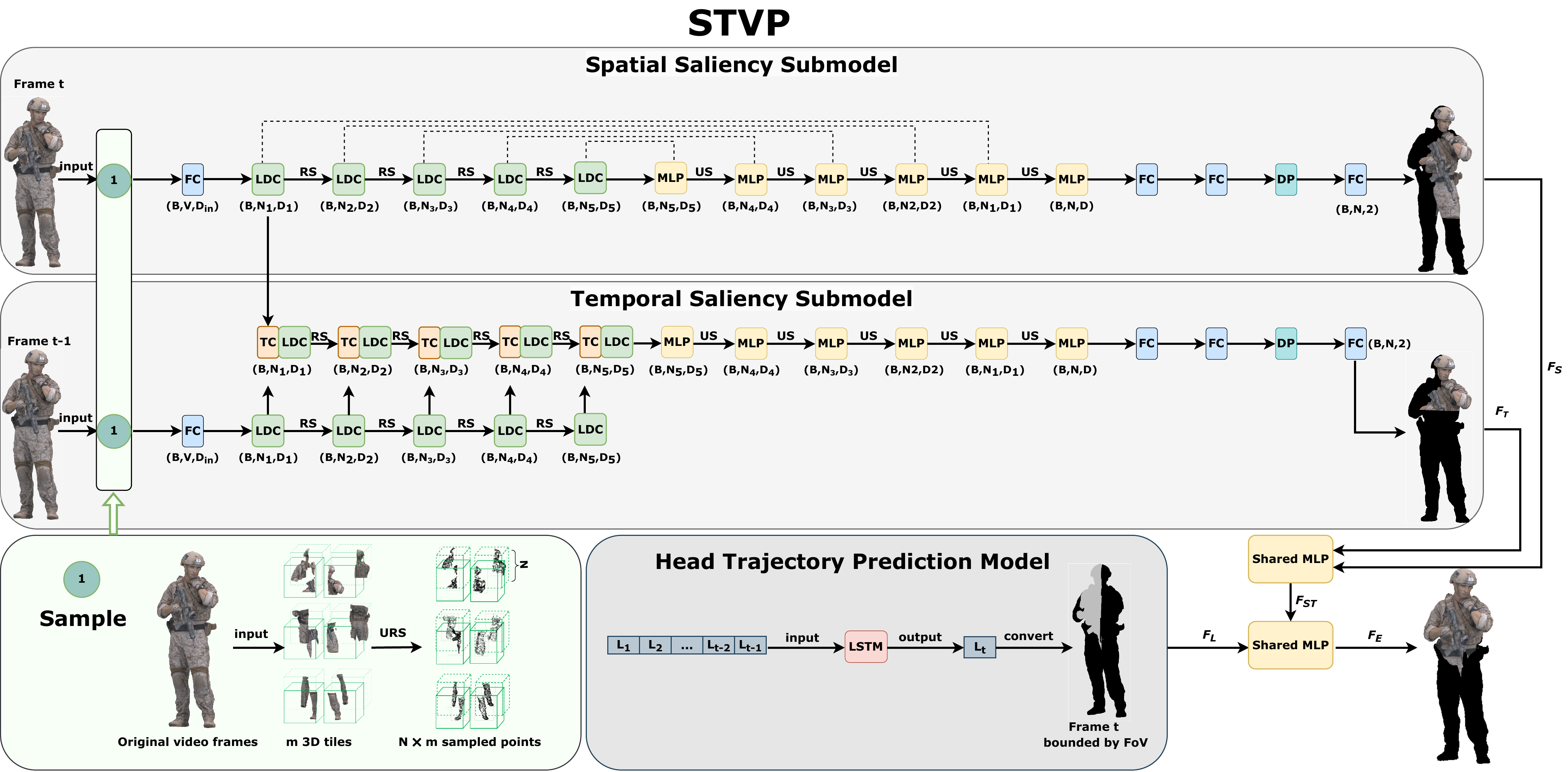}
    \caption{Overview of STVP.
    FC: Fully Connected Layer. LDC: Local Discrepancy Catcher. URS: Uniform-Random Sampling. DP: Dropout Layer. TC: Temporal contrast layer. MLP: Multi-Layer Perception. US: Up-Sampling. The $c$-th LDC module takes an  $B * N_c * D_c$ matrix as input that is from $B$ batches, $N_c$ points with $D_c$-dim point feature.
    $F_S$ and $F_T$ are the spatial and temporal salient features, respectively. $F_L$ is the predicted trajectory of the LSTM module, and is adaptively fused with $F_S$ and $F_T$ to generate the final FoV estimation, $F_E$. 
    The consecutive frames are first reduced numbers by URS, meanwhile the spatial-temporal information of the input data is retained as much as possible, then the spatial and temporal saliency features are learned in the spatial and temporal saliency sub-models, respectively. The user's historical head state is fed into the LSTM network to analyze and predict the user's head state when viewing the next frame.
    }
    \label{fig_network_overview}
\end{figure*}

Video saliency detection is more intricate, involving both spatial and temporal information, and has garnered significant research attention.
W. Wang et al. \cite{8578612} introduce an attentive CNN-LSTM architecture, which encodes static attention into dynamic saliency representation, leveraging both static and dynamic fixation data. M. Paul et al. \cite{8374803} propose a novel video summarization framework based on eye tracker data, calculating motion saliency scores by considering the distance between the viewer's current and previous focus, thus aiding video summarization. However, this approach focuses on extracting salient frames rather than the regions within those frames.

Building on these insights, we recognize the need to capture both static and dynamic saliency cues for point cloud video saliency detection.
K. Zhang et al. \cite{8543830} present a spatio-temporal dual-stream network that uses a cascaded architecture of deep 2D-CNN and shallow 3D-CNN for temporal saliency. While it excels in feature extraction, it lacks exploration of feature fusion.
Z. Wu et al. \cite{8474977} design three deep fusion modules-summation, maximization, and multiplication-to effectively utilize spatiotemporal features. However, these operations cannot discern the varying impacts of temporal and spatial saliency features on visual saliency detection in distinct scenarios.
Other researchers propose fusing features from different layers or points to obtain more important features \cite{9932589,10.5555/3295222.3295263,groh2018flex1,8578570}.

Moreover, traditional CNN networks are ineffective for processing unordered inputs like point cloud data. While treating points in 3D space as a volumetric grid offers an alternative, it's only practical for sparse point clouds, and CNNs remain computationally intensive for high-resolution ones.
Therefore, towards point-cloud video streaming, some studies employ scene flow methods to calculate motion vectors for all points in a frame, capturing temporal information across multiple frames \cite{8953933, liu2019flownet3d, 9435105}. Nevertheless, with point count in each frame reaching approximately $10^6$ or more, such an approach require substantial computational resources.

This paper presents an innovative network for saliency detection in point cloud videos, designed to handle the unordered nature of the data. Our approach focuses on capturing both static and dynamic saliency cues, making it highly effective for effective point cloud video saliency detection.

\subsection{Point Cloud Sampling}

Point cloud sampling is commonly used to reduce the computational complexity in point cloud video analysis. 
Heuristic sampling methods can be categorized into into five categories: farthest point sampling (FPS), inverse density importance sampling (IDIS), voxel sampling (VS), random sampling (RS), and geometric sampling (GS). GS considers the curvature of points, RS employs random probability for sampling, and IDIS samples based on inverse density. However, GS, RS, and IDIS cannot consistently preserve inter-frame dynamic components in every sampling outcome.

On the other hand, FPS and VS, due to their uniform spatial distribution, are capable of retaining key information in point clouds, including inter-frame dynamic regions.
Next, we analyze the time complexity of FPS and VS.

Assuming the original point cloud consists of approximately $N$ point, and our goal is to sample $V$ points, the time complexity of FPS, which selects each $p^v$ in the sampling point set $\{p^1, \ldots, p^v, \ldots, p^V\}$ as the farthest point from the first $v-1$ points, is $O(N^2)$. For VS, which divides the point cloud into $V$ voxels of equal volume (with time complexity $O(N)$) and computes the centroid of the points within each voxel to replace all the points in the voxel (with time complexity $O(V*\frac{N}{V})=O(N)$), the final time complexity is $O(N)$.

In summary, FPS has higher computational complexity, making it unsuitable for point cloud videos with large amounts of data. In Section \ref{sec:saliency}, we provide a detailed introduction to the URS sampling process and compare the performance of all sampling methods in Sec.\ref{sec:experimental}.

\section{Overview of the Proposed Viewport Prediction}\label{sec:overview}

When attracted by conspicuous geometric shapes, vivid colors, or moving objects, a user will shift their view towards the specific region by gradually altering their trajectory. Based on this intuition, we propose STVP, which consists of two main models: a saliency detection model and a user head trajectory prediction model. The saliency detection model is further divided into temporal saliency detection and spatial saliency detection to identify spatial and temporal salient regions, respectively. The detailed structure of STVP is shown in Fig. \ref{fig_network_overview}. Below, we describe the general principles of the saliency model and the trajectory prediction model.

\textbf{Prerequisites:} To ensure the performance of saliency detection, we need to make the sampling results of adjacent frames similar in the temporal dimension to preserve temporal information, while ensuring that the sampling points are aggregated in local space to preserve spatial information.
Furthermore, due to the large volume of point cloud video data, an effective sampling method is essential. To address this issue, we propose a saliency detection model that incorporates a block-based URS unit.

\textbf{Input:} We assume there are $T$ frames in a point cloud video, and $N^t_{in}$ points in frame $t$, each representing a 3D scene. 
Given consecutive 3D point cloud frames, we extract spatial features from the current frame and capture temporal features between consecutive frames in the spatial-temporal saliency detection sub-model.

In addition, the user's head state records $(L_1, L_2, \ldots, L_l, \ldots, L_{t-1})$ before time $t$ are fed to the trajectory prediction model to predict the head state $L_t$ of the current frame $t$, where the user's head state $L_l$ at time $l$ is consists of the head coordinates $(X_{l}, Y_{l}, Z_{l})$ and the Euler angles of the head rotation $(\alpha_{l}, \beta_{l}, \gamma_{l})$, i.e., $L_l = \{X_{l}, Y_{l}, Z_{l}, \alpha_{l}, \beta_{l}, \gamma_{l}\}$, $L_l \in \mathbf{R}^6$.

\textbf{Saliency detection:}
To reduce computational power consumption and enhance processing efficiency, the original video frames are first sampled through the URS unit before being input to the temporal and spatial saliency detection sub-models in parallel.
Then we propose an LDC module that catches local color and coordinate discrepancies to learn spatial saliency features $F_S$, along with a 3D-block based TC layer to extract temporal saliency features $F_T$. The details of the saliency detection model are provided in Section \ref{sec:saliency}.

\textbf{Head trajectory prediction and feature fusion:}
To predict the user's head state $L_{t}$ for frame $t$, we use LSTM to analyze the patterns in historical trajectories.

The predicted result $L_{t}$ is then transformed into trajectory features $F_L$ by assigning different color attributes to points inside and outside the viewport.

To achieve accurate viewport prediction, we combine the results of saliency detection with those of trajectory prediction through an adaptive fusion process. The adaptive fusion of spatial-temporal saliency features $F_S$, $F_T$ and trajectory features $F_L$ utilizes an attention mechanism to generate viewport prediction estimates $F_E$. Further details on head trajectory prediction and feature fusion are provided in Section \ref{sec:LSTM}.

\begin{figure*}[!htb]
    \centering
    \includegraphics[width=0.6\textwidth]{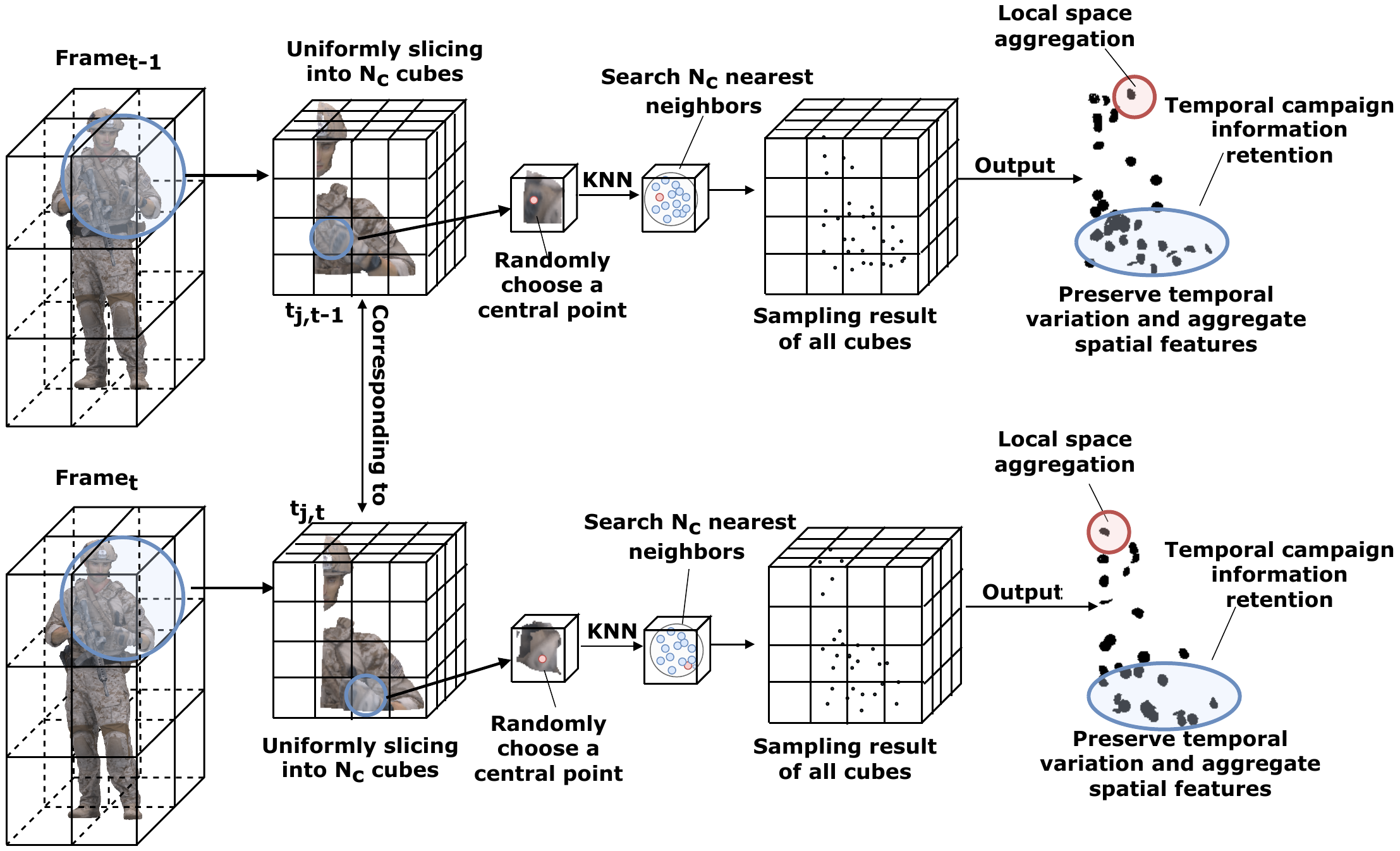}
    \caption{Detailed process of URS. The most distinct point cloud motion between two mutually mapped \textcolor{black}{3D blocks} $t_{k,t-1}$ and $t_{k,t}$ is marked by blue circles. The red points represent randomly selected central points within the block. The effect of spatial and temporal information retention is circled on the output.}
    \label{fig_URS}
\end{figure*}

\section{Spatial-temporal saliency detection model }\label{sec:saliency}

In this section, we will introduce the URS unit, the spatial saliency detection submodel and the temporal saliency detection submodel.

\subsection{Uniform-Random Sampling}\label{section:sample}

Sampling to reduce the number of points in a point cloud video provides several benefits, including lower computational and communication costs. The URS method is proposed for effective, uniformly spaced sampling that preservs temporal information, while KNN is used to aggregate spatial information from neighboring points.

Assuming each frame contails $N$ points, the frame is divided into $M$ \textcolor{black}{3D blocks} along the spatial dimension for transmission. Let $t_{j,t}$ denote the $j$-th \textcolor{black}{3D block} in frame $t$, with the total number of sampling points denoted as $V$. Additionally, let $D_{j,t}$ represent the sampled data of the \textcolor{black}{3D block} $t_{j,t}$ in frame $t$.

As shown in Fig. \ref{fig_URS}, we establish a mapping relationship between the \textcolor{black}{3D blocks}, $t_{j,t}$ and $t_{j,t-1}$, as they occupy similar spatial locations in their respective frames. This mapping is crucial for the TC layer to analyze changes in global information at corresponding locations.

To preserve temporal information, we evenly distribute sampling points within the mapped \textcolor{black}{3D blocks} to ensure consistent sampling results.
For more efficient sampling, we further divide each \textcolor{black}{3D block} into $N_c$ (ranging from 10 to $10^2$) smaller blocks and randomly select the central points. Around these central points, we search for $N_n$ neighbors as sampling points using KNN to aggregate local spatial information, where $N_n=\frac{V}{m*N_c}$, and $\frac{V}{m}$ represents the total number of sampling points in each \textcolor{black}{3D block}. By employing the URS method, we can efficiently select evenly spaced sampling points to preserve temporal information. Furthermore, the URS sampling process utilizes random sampling, further enhancing sampling efficiency.
The overall time complexity of URS is $O(N)$, meaning that both VS and URS have lower time complexity than FPS. To further compare the efficiency of VS and URS, we perform comprehensive experiments in Sec.\ref{sec:experimental}.

The TC layer implementation necessitates that sampled data with similar spatial locations maintain a close mapping relationship. The degree of this closeness can be quantified by the proposed inter-frame mapping strength in Sec.\ref{sec:experimental}, which evaluates the extent of temporal information retention.

\subsection{Spatial Saliency Detection Sub-model}\label{sec:spa-saliency}

The sampling result $\{D_{1,t},D_{2,t},... ,D_{j,t},... ,D_{m,t}\}$ of frame $t$ obtained through URS, is passed to the Fully Connected (FC) layer for feature extraction, the LDC module for spatial encoding, and the RS layer for down-sampling.

Given the coordinate attributes $p_{i,t}=\{x_{i,t},y_{i,t},z_{i,t}\}$ and color attributes $a_{i,t}=\{r_{i,t},g_{i,t},b_{i,t} \}$ for each point $i$ in $D_{j,t}$, we first extract the initial feature $f_{i,t}$ of point $i$ using the FC layer. The feature extraction process can be expressed as follows:
\begin{equation}
   f_{i,t}=FC(p_{i,t} \oplus a_{i,t})
\end{equation}
where $\oplus$ denotes concatenation. Finally, the sampled data $D_{j,t}$ passes through the FC layer to obtain the initial features $F_{j,t}$. Then we use the LDC module to fully integrate local coordinate and color discrepancies (i.e., merge spatial and color discrepancies into a unified representation). 
The key components of the LDC module are divided into three parts: \textit{Neighborhood Coding, Attention Pooling}, and \textit{Dilated Residual Blocks}.

\textbf{Neighborhood Coding.} 
This Neighborhood Coding unit explicitly incorporates the coordinate and color discrepancies between each point $i$ in $D_{j,t}$ and its neighboring points.
This ensures that the corresponding point features remain aware of their relative spatial differences, thereby enhancing the network's ability to effectively learn spatial salient features.

For the coordinate and color attributes $p_{i,t}^k$ and $a_{i,t}^k$ of a neighboring point $k$ around the point $i$, where $a_{i,t}^k=\{ r_{i,t}^k,g_{i,t}^k, b_{i,t}^k\}$ denotes the color attributes of point $k$, 
we explicitly encode the relative discrepancies as follows:
\begin{equation}
    d_{i,t}=0.299*r_{i,t}+0.587*g_{i,t}+0.114*b_{i,t}
\end{equation}
\begin{equation}
\begin{split}
    dp_{i,t}^k=& MLP[ p_{i,t}\oplus p_{i,t}^k \oplus (p_{i,t}- p_{i,t}^k) \oplus \Vert p_{i,t}- p_{i,t}^k \Vert\\
    &\oplus d_{i,t}
    \oplus d_{i,t}^k \oplus (d_{i,t}- d_{i,t}^k) \oplus \Vert d_{i,t}- d_{i,t}^k \Vert ]
\end{split}
\end{equation}
\begin{equation}
   \hat{f_{i,t}^k} =dp_{i,t}^k \oplus f_{i,t}^k
\end{equation}
where the constants $0.299$, $0.587$, and $0.114$ are used in the luminance algorithm \cite{1987Digital} to convert RGB values to grayscale $d_{i,t}$, reflecting the human eye's varying sensitivity to red, green, and blue. $\Vert {\cdot} \Vert$ represents the relative Euclidean distance in the coordinate space.

The discrepancy encoding $dp_{i,t}^k$  between point $i$ and its neighboring point $k$ is combined with the initial feature $f_{i,t}^k$ to generate the enhanced point feature $\hat{f_{i,t}^k}$.
Then, a set of enhanced features for point $i$, encoded with all $K$ neighboring points, is obtained as $A_{i,t}=\{\hat{f_{i,t}^1},\hat{f_{i,t}^2},\hat{f_{i,t}^3}...\hat{f_{i,t}^k},..,\hat{f_{i,t}^K}\}$ through neighborhood coding.

\textbf{Attention Pooling.}

This unit aggregates each feature $\hat{f_{i,t}^k}$ in the set of enhanced features $A_{i,t}$ into the spatial discrepancy feature $\hat{f_{i,t}}$ for point $i$. Thus, each point $i$ contains the features of its neighboring points, which compensates for the reduction in both temporal and spatial information caused by replacing URS with RS as the downsampling technique in the subsequent network training process.

For each enhanced feature $\hat{f_{i,t}^k}$, we first employ a shared function to compute its attention score $S_{i,t}^k$.
Next, we apply the \textit{Softmax} function  to normalize the attention score $S_{i,t}^k$, and use the normalized score to weight the enhanced feature, thereby obtaining the spatial discrepancy feature $\hat{f_{i,t}}$ as follows:

\begin{equation}\label{eq:score}
    S_{i,t}^k= \gamma(\hat{f_{i,t}^k},W)
\end{equation}
\begin{equation}\label{eq:weight}
    \hat{f_{i,t}}= \sum_{i=1}^{K}[\hat{f_{i,t}^k} *\sigma(S_{i,t}^k)]
\end{equation}
where $W$ is the shared weight, $\sigma ()$ is \textit{Softmax} function, and $\gamma()$ means Shared MLP function. 

\textbf{Dilated Residual Blocks.}

Neighborhood coding and attention pooling enable each sampled point to be spatially linked to $K$ neighbors. To expand the receptive field of each point, we perform neighborhood coding and attention pooling twice within dilated residual blocks, allowing information to be received from up to $K^2$ neighbors.

Based on the encoding process of the LDC module described above, the procedure of passing the initial features $F_{j,t}$ of the sampled data $D_{j,t}$ through the $c$-th LDC module and RS layer can be simplified as follows:
\begin{equation}\label{eq:cthLDCRS}
  \hat{F_{j,t}^{c}}=R_c\{LDC_c[\hat{{F_{j,t}^{c-1}}}],{\theta}_u]\}
\end{equation}
\begin{equation}\label{eq:initial}
  \hat{F_{j,t}^0}=L_1(F_{j,t})
\end{equation}
where $\hat{F_{j,t}^{c-1}}$ denotes the encoded features of $D_{j,t}$ processed by the $(c-1)$-th  LDC and RS layers, which are then passed as input to $c$-th LDC module. The input to the first LDC module, $\hat{F_{j,t}^0}$, is initialized according to Eq. \ref{eq:initial}. $LDC_c()$  represents the $c$-th LDC module, ${\theta}_u$ denotes the learnable weight, and $R_c$ represents the $c$-th RS layer.

The encoded features are decoded by passing them through MLP layers, Up-Sampling (US) layers, and FC layers. To further prevent overfitting, a Dropout (DP) layer is incorporated. The output of this decoding process is a set of spatial saliency features, represented by $F_S$.

\subsection{Temporal Saliency Detection Sub-model}\label{sec:tem-saliency}

Based on the mapping relationship between $D_{j,t}$ and $D_{j,t-1}$ extracted from frames $t$ and $t-1$, we propose a TC module to capture temporal information by comparing the feature differences between them. The feature difference reflects the degree of variation from $D_{j,t}$ to $D_{j,t-1}$. 
The general structure of the TC module is shown in Fig. \ref{fig_temporal} and is described as follows: 

\textbf{Maxpooling.} First, compute the global features of the two corresponding 3D blocks, $D_{j,t}$ and $D_{j,t-1}$, separately:

\begin{equation} \label{eq:maxpool1}
    Q_t^c=M(TC_{t}^c)
\end{equation}
\begin{equation}\label{eq:maxpool2}
    Q_{t-1}^c=M(TC_{t-1}^c)
\end{equation}
\noindent where $TC_{t-1}^c$ and $TC_{t}^c$ denote the inputs of the $c$-th TC from the \textcolor{black}{3D block} $D_{j,t}$ and \textcolor{black}{3D block} $D_{j,t-1}$, respectively. The initial values, $TC_{t-1}^1$ and  $TC_{t}^1$, are initialized as the first encoded features, $\hat{F_{j,t-1}^1}$ and $\hat{F_{j,t}^1}$, respectively. $Q_t^c$ and $Q_{t-1}^c$ refer to the global features of $TC_{t-1}^c$ and $TC_{t}^c$, respectively, computed through the \textit{Max pooling} function $M()$.

\textbf{Comparing Feature Similarity.} The corresponding global features are compared for similarity, and the similarity score is calculated and converted into saliency intensity. It is defined as follows:
\begin{equation}\label{eq:shared_mlp}
    S_{sim}=\gamma(Q_t^c \oplus  Q_{t-1}^c)
\end{equation}
\begin{equation} \label{eq:G}
    O_s=G(S_{sim})
\end{equation}
where $S_{sim}$ denotes the similarity score calculated between the corresponding global features, $Q_t^c$ and $Q_{t-1}^c$. $O_s$ represents the intensity of temporal saliency obtained by applying the temporal saliency operator, $G()$, to the similarity score, $S_{sim}$. The operator is defined as $G(\cdot)= \frac{1}{1 +\exp()} + 1$.
A lower degree of variation and smaller $O_s$ are expected when the similarity score, $S_{sim}$, between the global features $Q_t^c$ and $Q_{t-1}^c$ is higher.

\textbf{Weighting Features.} The input feature is weighted by the saliency intensity, as defined follows:
\begin{equation} \label{eq:result}
    C_t^c=O_s*TC_{t}^c
\end{equation}
\begin{equation}\label{eq:cthLDCRS1}
  \hat{TC_{t}^{c+1}}=R_c\{LDC_c[\hat{{TC_{t}^{c}}}],{\theta}_u]\}
\end{equation}
where the temporal change features, $C_t^c$, are obtained by weighting the input features, $TC_{t}^c$, from frame $t$, with the temporal saliency intensity. The minimum saliency intensity is set to 1. This means that if the saliency intensity exceeds 1, it will be superimposed after 5  TC layers. Otherwise, it will remain 1, with no change to the input features.
After applying LDC and RS, the resulting features are passed as input to the next TC module.


\begin{figure*}[!htb]
    \centering
    \includegraphics[width=0.8\linewidth]{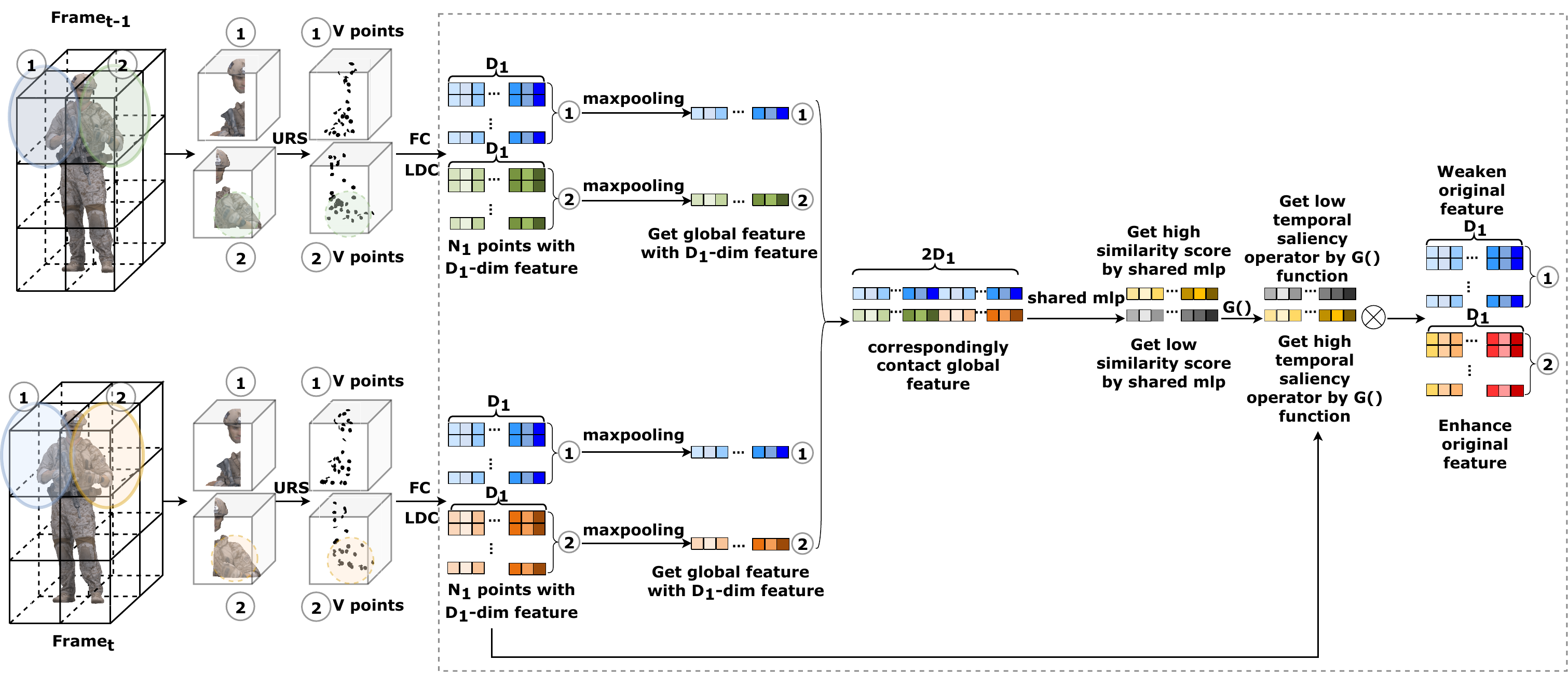}
    \caption{The structure of the temporal contrast layer (inside the dashed box). We choose two pairs of \textcolor{black}{3D blocks} mapped to each other in consecutive frame pairs to demonstrate the principle of the TC layer, where the point clouds within one pair of mapped \textcolor{black}{3D blocks} are relatively similar and the point clouds within the other pair are clearly moving.}
    \label{fig_temporal}
\end{figure*}

With multiple TC modules, we can efficiently capture the temporal information between the mapping \textcolor{black}{3D blocks}, and subsequently obtain the temporal salient features, $F_T$, by decoding through MLP, US, and FC layers.

\section{User's Head Trajectory Prediction Module and Feature Fusion Unit}\label{sec:LSTM}

This section presents the LSTM-based module for predicting head trajectory, as well as the feature fusion unit that utilizes attention mechanisms.

Since LSTM has the capacity to learn temporal dependencies between observations, we use it to predict the user's head state at the next time step $t$, based on the previous head position.
The state update process of the user's head state $L_l$ (where $l$ represents the time index) through the LSTM cell is shown in the following equation: 
\begin{equation}
        h_l=f(h_{l-1},L_l)
\end{equation}
where $h_{l}$ is the hidden state of the LSTM cell structure at $l$-th moments. 
$f()$ is the nonlinear activation function of the LSTM cell. Each LSTM cell consists of a forget gate $f_l$, a input gate $i_l$, a output gate $o_l$, and a cell state $C_{l}$. The specific  encoding process of LSTM cells is as follows:

\begin{equation}
      f_l=\sigma(W_f[h_{l-1},L_l]+b_f)
   \end{equation}    
    \begin{equation}
      i_l=\sigma(W_i[h_{l-1},L_l]+b_i)
    \end{equation} 
     \begin{equation}
      o_l=\sigma(W_o[h_{l-1},L_l]+b_o)
    \end{equation} 
       \begin{equation}
      \Tilde{C_l}=tanh(W_c[h_{l-1},L_l]+b_c)
       \end{equation}
 \begin{equation}
      C_l=f_l*c_{l-1}+i_l*\Tilde{C_l}
  \end{equation}
 \begin{equation}
      h_l=o_l*tanh(C_l)
  \end{equation}  
  
where $[h_{l-1},L_l]$ represents the connection vector between the hidden state $h_{l-1}$ and the input $L_l$. 
$W_f$, $W_i$, $W_o$, $W_c$ are the learnable weights in the network and
 $b_f,b_i,b_o,b_c$ are the learnable biases in the network.
$\sigma()$ denotes the \textit{Sigmoid} activation function. Then, the hidden state $h_l$  and the next head state $L_{l+1}$ are passed to the next LSTM cell to learn the temporal dependence.

At the last LSTM cell, the predicted result of the user's head state , $L_t$, is obtained. 
The predicted user's head state is mapped to the user's viewport, and the current frame $t$ is partitioned into two regions based on it relative location to the viewport, as illustrated in Fig. \ref{fig_network_overview}. An RGB color scheme is assigned to the points inside and outside the viewport, where points inside the viewport are colored white (RGB values of 255,255,255), and points outside the FoV are colored black (RGB values of 0,0,0).
Subsequently, the features from the current frame are extracted as trajectory features, $F_L$.

Then, we integrate spatial saliency features $F_S$ and temporal saliency features $F_T$ into the combined saliency features $F_{ST}$, which are then fused with trajectory features $F_L$ to obtain final predicted estimates, $F_t$.

Traditional feature aggregation methods, such as \textit{Maxpooling} and \textit{Averagepooling}, often lead to a significant loss of point information, as they aggregate features without considering their relationships. To mitigate this loss, 
we employ an attention mechanism to adaptively fuse the spatial-temporal saliency features of point cloud videos. The specific fusion process is as follows:

\textbf{Computing attention scores.} Given the spatial saliency features $F_S$ and the temporal saliency features $F_T$, we first apply a shared MLP followed by
\textit{Softmax} function to calculate the attention scores. This process is formally defined as follows:

\begin{equation}
    \centering
    S_S=\sigma[\gamma(F_S,W_1)]
\end{equation}
\begin{equation}
    S_T=\sigma[\gamma(F_T,W_2)]
\end{equation}
where $W_1,W_2$ is the learnable weight, $\sigma()$ is the \textit{Softmax} function, and $\gamma()$ is a proxy for shared MLP. 

\textbf{Weighting Features.} The learned attention scores can be interpreted as a mask for the automatically selecting the salient features. These features are weighted to produce the saliency feature $F_{ST}$, which is represented as follows:

\begin{equation}
    F_{ST}=S_S \odot F_S + S_T \odot F_T
\end{equation}
where $\odot$ denotes the element-wise multiplication.

Second, we use the same fusion approach to fuse salient features with trajectory features $F_L$:
\begin{equation}
    S_{ST}=\sigma[\gamma(F_{ST},W_3)]
\end{equation}
\begin{equation}
    S_L=\sigma[\gamma(F_L,W_4)]
\end{equation}
\begin{equation}
    F_E=S_{ST} \odot F_{ST} + S_L \odot F_L
\end{equation}
where $W_3,W_4$ are the learnable weight parameters, $\sigma$ denotes the \textit{Softmax} function, $\gamma$ represents a proxy for the shared MLP, and $\odot$ denotes the element-wise multiplication. Finally, the viewport prediction features $F_E$ are obtained, which divide the points into those within the viewport and those outside the viewport. 

\section{Experiment}
\label{sec:experimental}
In this section, we verify the effectiveness of our proposed prediction method through extensive experiments.

\subsection{Experimental Setup}

\subsubsection{Datasets} We select four point cloud video sequences from the 8iVFB dataset \cite{d20178i}, namely ``Soldier'', ``Redandblack'', ``Loot'', and ``Longdress'', as shown in Fig. \ref{fig_dataset}. These sequences provide comprehensive coverage of the full body of the portrait characters, allowing users to observe their movements. During the model training phase, we allocate 900, 150, and 150 frames for training, validation, and testing, respectively. We conduct experiments on three networks (STVP, RandLANet, BAAF-Net) using 12, 20, and 36 \textcolor{black}{3D blocks} ($M={12,20,36}$) to analyze the impact of varying  the number of \textcolor{black}{3D blocks} on viewport prediction.

\begin{figure}[!htb] 
    \centering
      \includegraphics[width=0.8\linewidth]{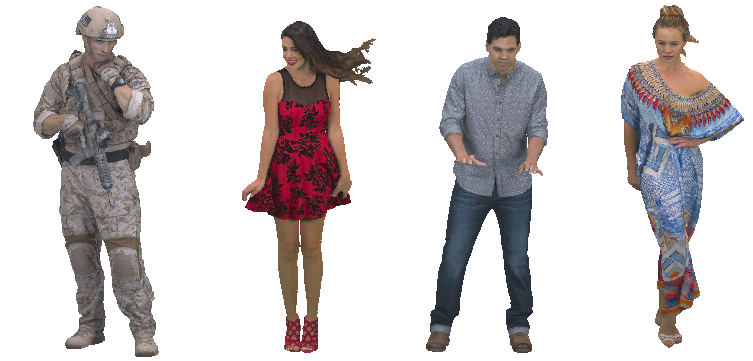} \label{data1}
       \caption{Point cloud video sequences: Soldier, Redandblack, Loot, Longdress.}
	  \label{fig_dataset}
\end{figure}

\begin{figure}[htb!]
    \centering

      \includegraphics[width=0.7\linewidth]{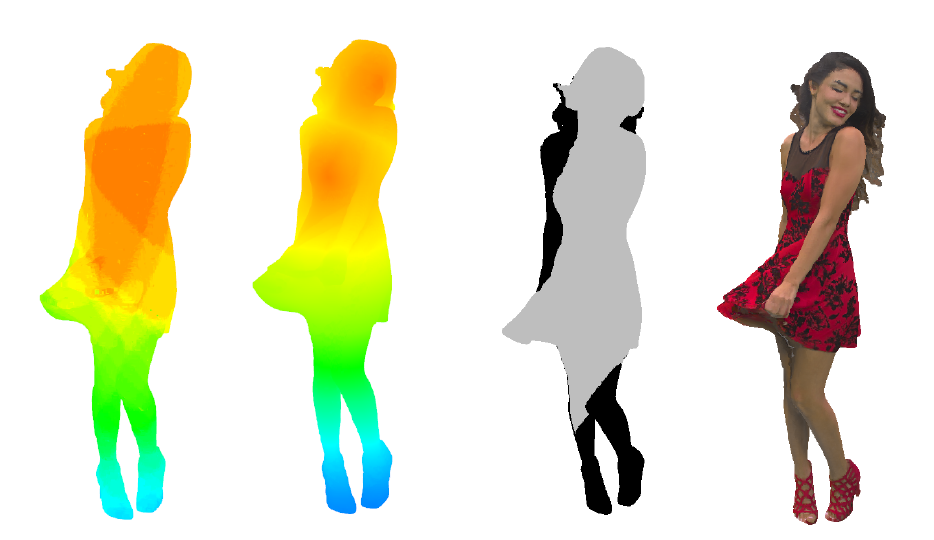}
      \hspace{0.5mm}
 
\caption{Illustration of the gaze heatmap, the saliency, the viewport predicted and the orgiginal video frame. The first figure represents the gaze heatmap of 40 volunteers, showing their visual attention distribution across the point cloud scene. The second figure displays the saliency map for the corresponding point cloud video frame. The third figure shows the FOV prediction results from our model, while the last figure presents the original point cloud video frame.}
	  \label{map}
\end{figure}

We recruit 40 student volunteers (25 males and 15 females) for our experiment to view point cloud videos using HMDs. The experiment requires participants to experience a 6DoF immersive environment and explore areas of interest in a manner they deem appropriate during the viewing process.
The head trajectories of all participants are meticulously recorded throughout the entire duration of the video. Fig. \ref{map} illustrates the viewpoint heatmaps of users as they view different frames of the point cloud video.

\subsubsection{Implementation} 
The STVP model has been implemented using TensorFlow, and the source code can be downloaded using the following link\footnote{Source code URL: https://github.com/Cedarleigh/STVP-main}.
Details of the network architecture are provided in \textcolor{black}{Appendix}. 
All experiments are conducted on our AI server equipped with four Nvidia Tesla V100 GPUs.

\subsubsection{Metrics}
The overall performance of STVP is evaluated using various metrics.

(1) Sampling performance: The performance of our URS method is evaluated by comparing memory consumption, time consumption, and inter-frame mapping intensity (IFMI) with those of other sampling methods. 

To evaluate the IFMI of different sampling methods, we propose a metric called Distance And Color Variation Values (DaCVV).
This metric calculates the coordinates and the degree of color difference between points $i$ and $l$ within the corresponding \textcolor{black}{3D blocks}, and is defined as:
\begin{equation}
   DaCVV=  \frac{\Vert p_{i,t}-p_{l,t-1} \Vert}{D_{max}}+\frac{\Vert a_{i,t}-a_{l,t-1} \Vert}{C_{max}}
\end{equation}
\noindent where the $D_{max}$ and $C_{max}$ are the max coordination and color distance of \textcolor{black}{3D block} $t_{k,t}$, respectively.

Thus, we can calculate the DaCVV between point $i$ from  $t_{k,t}$ and point $l$ from $t_{k,t-1}$.
If the DaCVV between two points is below a certain threshold, the two points are considered to be mapped to each other. 
We compute the ratio of all mapped points to all sampled points as the IFMI. We compare the IFMI of each sampling method for DaCVV thresholds of $\{0.1,0.2,0.3,0.4,0.5,0.6,0.7,0.8,0.9\}$.

(2) Saliency detection:
To assess the saliency detection performance of STVP, we compare its predicted saliency output with the ground truth and other competing models. The evaluation is conducted using four metrics: accuracy, precision, recall, and mean intersection over union (MIoU).

(3) Viewport prediction: We evaluate the performance of STVP in viewport prediction using the same metrics applied in saliency detection evaluation. In particular, we extend the MIoU metric into two categories for viewport prediction: point-level MIoU and block-level MIoU. Point-level MIoU measures the intersection-over-union ratio of point-level predicted results with the ground truth, while block-level MIoU measures the intersection-over-union ratio of block-level predicted results with the ground truth.

\subsubsection{Baselines} 
We compare STVP with the competitors in terms of sampling and saliency detection respectively.

(1) Baselines for sampling: The performance of URS is compared with five commonly used competitors: FPS \cite{10.5555/3295222.3295263}, GS, RS, IDIS \cite{groh2018flex1}, VS \cite{8578570}.

(2) Baselines for saliency detection: 
PointNet++ \cite{10.5555/3295222.3295263}, RandLANet\cite{9156466} and BAAF-Net\cite{9577557} have shown excellent performance in saliency detection. We compare the saliency detection performance of STVP saliency detection network (STVP-SD) with these models. Specifically, PointNet++ proposes a hierarchical ensemble abstraction method that takes a frame of point cloud video as input, performing hierarchical downsampling and neighborhood group coding processes on each frame. The extracted spatial features gradually capture more localized information. 
Inspired by PointNet++, RandLANet improves the group encoding layer by introducing a local feature aggregator to encode neighborhood coordination information, which allows for better learning of spatial features.
BAAF-Net takes a further step by proposing a bilateral context module to extend local coordinate encoding. This module first fuses the local semantic context information from the point cloud video frames with the local geometric context information, and then extracts local spatial features by adapting the geometric context to the local semantic context.

\subsubsection{Ablation setup}
In the experiments, we also conduct the following ablation studies for STVP. 

(1$\sim$2) RS++ and IDIS++: We replace the URS module in STVP with RS and IDIS, respectively, while keeping the STVP saliency extraction modules intact. These two variants are referred to as RS++ and IDIS++, respectively. This comparison is made to emphasize the contribution of the URS unit in retaining spatial-temporal information comparison.

(3) LFA++: To demonstrate the potential benefits of our proposed LDC, we replace the LDC module in STVP with the LFA module from RandLANet, which focuses solely on the geometric coordinate information of the neighborhood.
This comparison helps us better understand the performance of LDC, which leverages both neighboring coordinate information and color information to obtain more accurate local spatial saliency, as discussed in Sec. \ref{sec:saliency}.

(4) TSD--: 

We remove the temporal saliency detection (TSD--) module and obtain the viewport prediction results for this ablation study.
By comparing the prediction results of STVP and TSD--, we obtain the importance of the temporal saliency model in achieving accurate viewport prediction. 

\subsection{Experimental Results}

\textbf{The effect of the number of \textcolor{black}{3D blocks}:}
We hypothesize that the number of \textcolor{black}{3D blocks} affects viewport prediction. In our experiments, shown in Table \ref{tb:tile number}, varying the number of \textcolor{black}{3D blocks} (12, 20, and 36) resulted in a slight increase in MIoU at both the point and block-levels across different networks. However, the computation time to segment the entire point cloud video into individual \textcolor{black}{3D blocks} increases significantly with the block number. Therefore, in subsequent experiments, we set the number of \textcolor{black}{3D blocks} to 12.

\begin{table}[htb!]
\caption{Performance of the three networks with different numbers of 3D blocks.}
\scalebox{0.55}{
\begin{tabular}{c|cccc}
\hline
\multirow{2}{*}{Network}    & \multicolumn{1}{l|}{\multirow{2}{*}{The number of \textcolor{black}{3D blocks}}} & \multicolumn{2}{c|}{MIOU}                                                    & \multirow{2}{*}{slicing Time(s)} \\ \cline{3-4}
                            & \multicolumn{1}{l|}{}                                     & \multicolumn{1}{c|}{point-level MIoU} & \multicolumn{1}{c|}{\textcolor{black}{block-level MIoU}} &                                 \\ \hline
\multirow{3}{*}{STVP}       & \textbf{12}                                               & \textbf{82.09$\%$}                    & \textbf{88.10$\%$}                   & \textbf{300}                    \\ \cline{2-5} 
                            & \textbf{20}                                               & \textbf{82.67$\%$}                    & \textbf{89.03$\%$}                   & \textbf{420}                    \\ \cline{2-5} 
                            & \textbf{36}                                               & \textbf{83.19$\%$}                    & \textbf{89.90$\%$}                   & \textbf{930}                    \\ \hline
\multirow{3}{*}{RandLA-Net} & 12                                                        & 61.66$\%$                             & 74.32$\%$                            & 289                             \\ \cline{2-5} 
                            & 20                                                        & 62.36$\%$                             & 75.15$\%$                            & 315                             \\ \cline{2-5} 
                            & 36                                                        & 63.45$\%$                             & 75.99$\%$                            & 921                             \\ \hline
\multirow{3}{*}{BAAF-Net}   & 12                                                        & 64.66$\%$                             & 77.29$\%$                            & 275                             \\ \cline{2-5} 
                            & 20                                                        & 65.97$\%$                             & 78.06$\%$                            & 422                             \\ \cline{2-5} 
                            & 36                                                        & 66.03$\%$                             & 79.60$\%$                            & 913                             \\ \hline
\end{tabular}
\label{tb:tile number}
}
\end{table}

\begin{figure}[htb!]
    \centering
	  \subfloat[]{
   \includegraphics[width=0.48\linewidth]{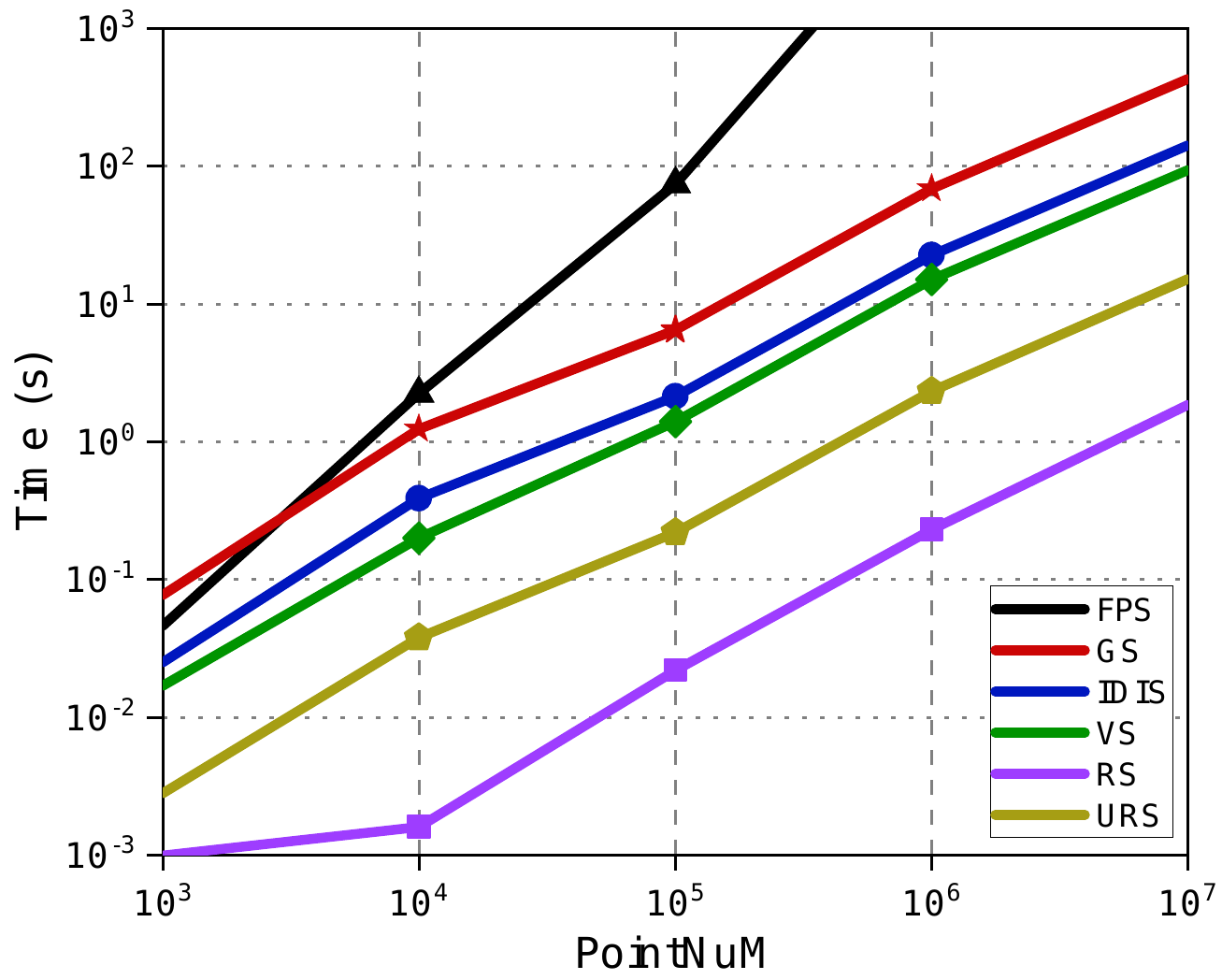} \label{sr1a}}
       \subfloat[]{  \includegraphics[width=0.48\linewidth]{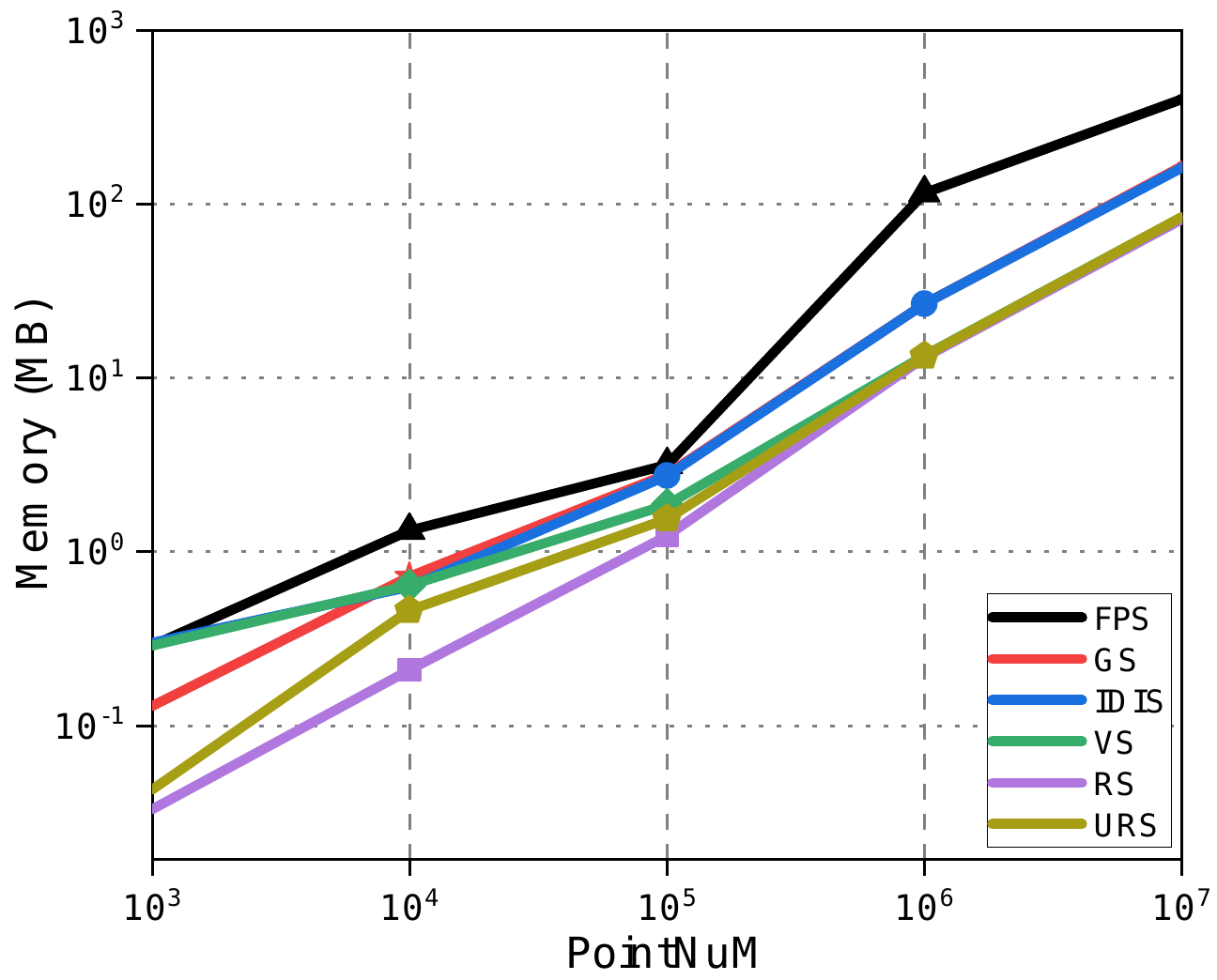} \label{sr1b}} 

    \subfloat[]{ \includegraphics[width=0.48\linewidth]{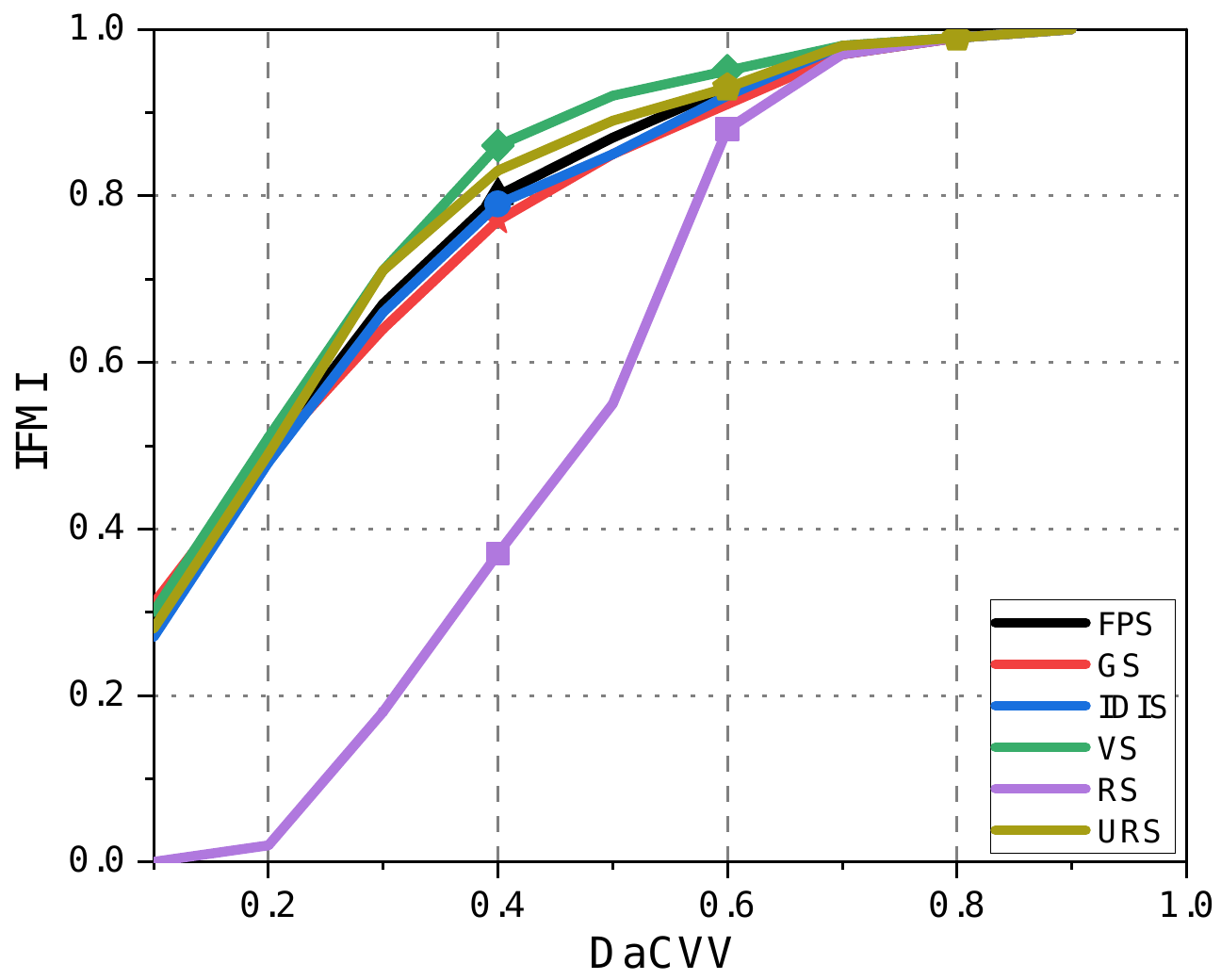}
    \label{sr1c}
    } 
       
	  \caption{Time consumption, memory consumption and IFMI value of different sampling approaches.}
	  \label{fig_sresult1} 
   
\end{figure}

\textbf{The efficiency of URS:}

To evaluate the efficiency of sampling, we select five levels of point cloud sizes: $\sim 10^3$ points, $\sim 10^4$ points, $\sim 10^5$ points,  $\sim 10^6$ points, and $\sim 10^7$ points. We downsample the point clouds to 1/4 of the original size using FPS, VS, IDIS, GS, RS, and URS, respectively. We measure the sampling time and memory consumption for each sampling method. Each experiment is run 50 times independently, and the results are averaged for the final outcome. Additionally, we present the sampling distribution for each method separately.

The time and memory consumption for all the sampling methods are shown in Fig.\ref{fig_sresult1}.
Experimental results indicate that the sampling time and memory consumption of URS are second only to RS, as URS requires slicing the point cloud into \textcolor{black}{3D blocks}. 
When the number of sampling points exceeds $10^5$, the memory consumption of URS, VS and RS is significantly lower than that of other methods. The confidence intervals for this sampling comparison experiment at the 95\% confidence level are shown in the Table \ref{tb:CI}.

\begin{figure}[htb!] 
    \centering
	  \subfloat[URS.]{
       \includegraphics[width=0.2\linewidth]{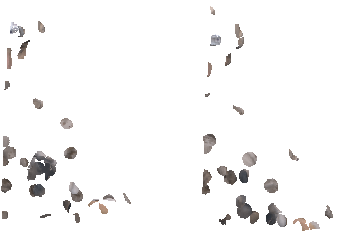} \label{sr2a}}
	  \subfloat[RS.]{
       \includegraphics[width=0.2\linewidth]{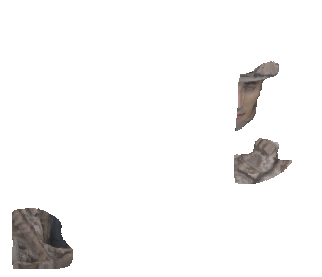} \label{sr2b}}\hspace{6mm}
       \subfloat[FPS.]{
       \includegraphics[width=0.2\linewidth]{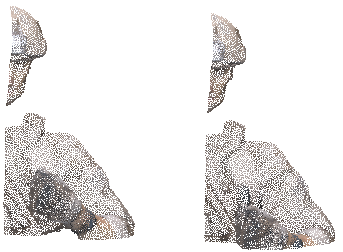} \label{sr2c}}
       
       \subfloat[GS.]{
       \includegraphics[width=0.2\linewidth]{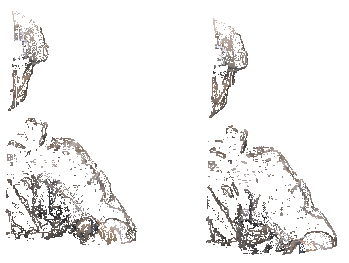} \label{sr2d}}\hspace{3mm} 
       \subfloat[VS.]{
       \includegraphics[width=0.2\linewidth]{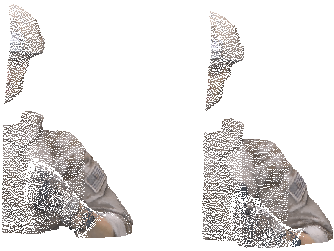} \label{sr2e}}\hspace{1mm}
       \subfloat[IDIS.]{
       \includegraphics[width=0.2\linewidth]{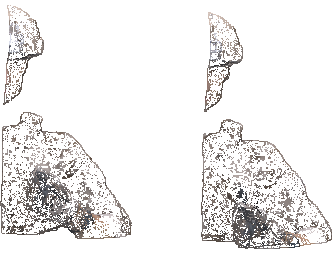} \label{sr2f}} 
	  \caption{Results of different sampling methods for consecutive frames.}
	  \label{fig_sresult2} 
\end{figure}

As shown in Fig. \ref{sr1c}, we compare the IFMI of the six sampling methods at the same DaCVV values. When the DaCVV values of URS are 0.3 and 0.8, its IFMI values are comparable to those of VS. In other cases, the IFMI values of URS are slightly lower than those of VS but higher than or equivalent to those of FPS, RS, GS, and IDIS across all DaCVV values. 
However, considering time consumption, as shown in Fig. \ref{sr1a}, when the number of points in the point cloud is around $10^7$, the time consumption of VS is 80 seconds higher than that of URS (with the number of points in a single frame ranges from 
$10^6$ to $10^7$, and a point cloud video sequence containing a total of 300 frames). URS achieves efficient sampling by trading off a slight loss in temporal information retention. 

To visually observe the sampling effect, we take the corresponding \textcolor{black}{3D blocks} $t_{k,t-1}$ and $t_{k,t}$ from consecutive frames $t-1$ and $t$ as sampling inputs, with the number of sampling points set to 12288, as shown in Fig.~\ref{fig_sresult2}. 
The sampling result shows that URS effectively retains both temporal and spatial information.
Although FPS and VS retain the temporal information between frames more effectively, they consume more sampling time and memory than URS and neglect spatial local information.
RS and GS preserve temporal information less effectively than URS.

\begin{table*}[]
\caption{The confidence intervals and means of time and memory consumption for each sampling method at 0.95 confidence level}
\label{tb:CI}
\begin{spacing}{1.2}
\scalebox{0.53}{
\begin{tabular}{|c|c|cc|c|c|cc|c|c|cc|c|}
\hline
Number of Points        & Sampling Methods      & \multicolumn{2}{c|}{\begin{tabular}[c]{@{}c@{}}Confidence interval\\ (Confience level 0.95)\end{tabular}} & Mean      & \multicolumn{1}{l|}{Sampling Methods} & \multicolumn{2}{c|}{\begin{tabular}[c]{@{}c@{}}Confidence interval\\ (Confience level 0.95)\end{tabular}} & Mean    & \multicolumn{1}{l|}{Sampling Methods} & \multicolumn{2}{c|}{\begin{tabular}[c]{@{}c@{}}Confidence interval\\ (Confience level 0.95)\end{tabular}} & Mean   \\ \hline
\multirow{2}{*}{$10^3$} & \multirow{10}{*}{FPS} & \multicolumn{1}{c|}{Time(s)}                           & {[}0.045,0.063{]}                                & 0.046     & \multirow{10}{*}{VS}                  & \multicolumn{1}{c|}{Time(s)}                             & {[}0.016,0.019{]}                              & 0.017   & \multirow{10}{*}{IDIS}                & \multicolumn{1}{c|}{Time(s)}                    & {[}0.024,0.026{]}                                       & 0.025  \\
                        &                       & \multicolumn{1}{c|}{Memory(MB)}                        & {[}0.26,0.31{]}                                  & 0.29      &                                       & \multicolumn{1}{c|}{Memory(MB)}                          & {[}0.27,0.31{]}                                & 0.29    &                                       & \multicolumn{1}{c|}{Memory(MB)}                 & {[}0.29,0.35{]}                                         & 0.30   \\ \cline{1-1} \cline{3-5} \cline{7-9} \cline{11-13} 
\multirow{2}{*}{$10^4$} &                       & \multicolumn{1}{c|}{Time(s)}                           & {[}2.21,2.32{]}                                  & 2.24      &                                       & \multicolumn{1}{c|}{Time(s)}                             & {[}0.20,0.21{]}                                & 0.20    &                                       & \multicolumn{1}{c|}{Time(s)}                    & {[}0.39,0.40{]}                                         & 0.39   \\
                        &                       & \multicolumn{1}{c|}{Memory(MB)}                        & {[}1.32,1.37{]}                                  & 1.33      &                                       & \multicolumn{1}{c|}{Memory(MB)}                          & {[}0.63,0.66{]}                                & 0.64    &                                       & \multicolumn{1}{c|}{Memory(MB)}                 & {[}0.57,0.75{]}                                         & 0.63   \\ \cline{1-1} \cline{3-5} \cline{7-9} \cline{11-13} 
\multirow{2}{*}{$10^5$} &                       & \multicolumn{1}{c|}{Time(s)}                           & {[}73.81,75.35{]}                                & 74.58     &                                       & \multicolumn{1}{c|}{Time(s)}                             & {[}1.38,1.41{]}                                & 1.40    &                                       & \multicolumn{1}{c|}{Time(s)}                    & {[}2.12,2.15{]}                                         & 2.14   \\
                        &                       & \multicolumn{1}{c|}{Memory(MB)}                        & {[}2.73,3.57{]}                                  & 3.16      &                                       & \multicolumn{1}{c|}{Memory(MB)}                          & {[}1.52,2.20{]}                                & 1.87    &                                       & \multicolumn{1}{c|}{Memory(MB)}                 & {[}2.64,2.85{]}                                         & 2.74   \\ \cline{1-1} \cline{3-5} \cline{7-9} \cline{11-13} 
\multirow{2}{*}{$10^6$} &                       & \multicolumn{1}{c|}{Time(s)}                           & {[}9917.09,9995.40{]}                            & 9956.25   &                                       & \multicolumn{1}{c|}{Time(s)}                             & {[}15.04,15.20{]}                              & 15.12   &                                       & \multicolumn{1}{c|}{Time(s)}                    & {[}22.64,22.81{]}                                       & 22.73  \\
                        &                       & \multicolumn{1}{c|}{Memory(MB)}                        & {[}114.40,116.83{]}                              & 115.61    &                                       & \multicolumn{1}{c|}{Memory(MB)}                          & {[}13.43,13.57{]}                              & 13.50   &                                       & \multicolumn{1}{c|}{Memory(MB)}                 & {[}26.52,26.86{]}                                       & 26.70  \\ \cline{1-1} \cline{3-5} \cline{7-9} \cline{11-13} 
\multirow{2}{*}{$10^7$} &                       & \multicolumn{1}{c|}{Time(s)}                           & {[}358771.65,366392.56{]}                        & 362582.11 &                                       & \multicolumn{1}{c|}{Time(s)}                             & {[}93.13,94.76{]}                              & 93.95   &                                       & \multicolumn{1}{c|}{Time(s)}                    & {[}141.93,143.42{]}                                     & 142.68 \\
                        &                       & \multicolumn{1}{c|}{Memory(MB)}                        & {[}240.38,269.67{]}                              & 254.36    &                                       & \multicolumn{1}{c|}{Memory(MB)}                          & {[}83.99,84.73{]}                              & 84.36   &                                       & \multicolumn{1}{c|}{Memory(MB)}                 & \multicolumn{1}{l|}{{[}161.44,162.82{]}}                & 162.13 \\ \hline
\multirow{2}{*}{$10^3$} & \multirow{10}{*}{GS}  & \multicolumn{1}{c|}{Time(s)}                           & {[}0.074,0.079{]}                                & 0.077     & \multirow{10}{*}{RS}                  & \multicolumn{1}{c|}{Time(s)}                             & {[}0.00097,0.00099{]}                          & 0.00097 & \multirow{10}{*}{URS}                 & \multicolumn{1}{c|}{Time(s)}                    & {[}0.0028,0.0029{]}                                     & 0.0028 \\
                        &                       & \multicolumn{1}{c|}{Memory(MB)}                        & {[}0.11,0.13{]}                                  & 0.13      &                                       & \multicolumn{1}{c|}{Memory(MB)}                          & {[}0.026,0.039{]}                              & 0.033   &                                       & \multicolumn{1}{c|}{Memory(MB)}                 & {[}0.026,0.045{]}                                       & 0.043  \\ \cline{1-1} \cline{3-5} \cline{7-9} \cline{11-13} 
\multirow{2}{*}{$10^4$} &                       & \multicolumn{1}{c|}{Time(s)}                           & {[}1.22,1.26{]}                                  & 1.24      &                                       & \multicolumn{1}{c|}{Time(s)}                             & {[}0.0012,0.0019{]}                            & 0.0016  &                                       & \multicolumn{1}{c|}{Time(s)}                    & {[}0.038,0.039{]}                                       & 0.038  \\
                        &                       & \multicolumn{1}{c|}{Memory(MB)}                        & {[}0.62,0.82{]}                                  & 0.72      &                                       & \multicolumn{1}{c|}{Memory(MB)}                          & {[}0.18,0.24{]}                                & 0.21    &                                       & \multicolumn{1}{c|}{Memory(MB)}                 & {[}0.37,0.55{]}                                         & 0.46   \\ \cline{1-1} \cline{3-5} \cline{7-9} \cline{11-13} 
\multirow{2}{*}{$10^5$} &                       & \multicolumn{1}{c|}{Time(s)}                           & {[}6.45,6.57{]}                                  & 6.51      &                                       & \multicolumn{1}{c|}{Time(s)}                             & {[}0.021,0.023{]}                              & 0.022   &                                       & \multicolumn{1}{c|}{Time(s)}                    & {[}0.21,0.23{]}                                         & 0.22   \\
                        &                       & \multicolumn{1}{c|}{Memory(MB)}                        & {[}2.68,2.89{]}                                  & 2.79      &                                       & \multicolumn{1}{c|}{Memory(MB)}                          & {[}1.47,1.63{]}                                & 1.10    &                                       & \multicolumn{1}{c|}{Memory(MB)}                 & {[}1.17,1.31{]}                                         & 1.24   \\ \cline{1-1} \cline{3-5} \cline{7-9} \cline{11-13} 
\multirow{2}{*}{$10^6$} &                       & \multicolumn{1}{c|}{Time(s)}                           & {[}68.77,69.65{]}                                & 69.21     &                                       & \multicolumn{1}{c|}{Time(s)}                             & {[}0.23,0.24{]}                                & 0.23    &                                       & \multicolumn{1}{c|}{Time(s)}                    & {[}2.33,2.34{]}                                         & 2.34   \\
                        &                       & \multicolumn{1}{c|}{Memory(MB)}                        & {[}26.54,26.71{]}                                & 26.62     &                                       & \multicolumn{1}{c|}{Memory(MB)}                          & {[}13.07,13.14{]}                              & 13.11   &                                       & \multicolumn{1}{c|}{Memory(MB)}                 & {[}13.35,13.41{]}                                       & 13.38  \\ \cline{1-1} \cline{3-5} \cline{7-9} \cline{11-13} 
\multirow{2}{*}{$10^7$} &                       & \multicolumn{1}{c|}{Time(s)}                           & {[}430.93,438.10{]}                              & 434.52    &                                       & \multicolumn{1}{c|}{Time(s)}                             & {[}1.85,1.88{]}                                & 1.86    &                                       & \multicolumn{1}{c|}{Time(s)}                    & {[}15.17,15.39{]}                                       & 15.28  \\
                        &                       & \multicolumn{1}{c|}{Memory(MB)}                        & {[}163.86,165.19{]}                              & 164.53    &                                       & \multicolumn{1}{c|}{Memory(MB)}                          & {[}81.29,81.75{]}                              & 81.52   &                                       & \multicolumn{1}{c|}{Memory(MB)}                 & {[}83.90,84.51{]}                                       & 84.21  \\ \hline
\end{tabular}
}
\end{spacing}
\end{table*}

\begin{table}[!htb]
\caption{Performance comparison of different networks.}
\centering
\label{tb_result}
\scalebox{0.6}{
\begin{tabular}{cllccccc}
\specialrule{0em}{2pt}{2pt}
\hline
\specialrule{0em}{2pt}{2pt}
\multicolumn{3}{c}{} &
  \multicolumn{2}{c}{\textbf{MIoU}} &
  \multirow{2}{*}{\textbf{Accuracy}} &
  \multirow{2}{*}{\textbf{Precision}} &
  \multirow{2}{*}{\textbf{Recall}} \\ \specialrule{0em}{2pt}{2pt} \cline{1-5} \specialrule{0em}{2pt}{2pt}
\multicolumn{3}{c}{} &
  \multicolumn{1}{l|}{\textbf{Point-level MIoU}} &
  \multicolumn{1}{l}{\textbf{block-level MIoU}} &
   &
   &
   \\\specialrule{0em}{2pt}{2pt} \hline \specialrule{0em}{2pt}{2pt}
\multicolumn{3}{c}{PointNet++} &
  52.22$\%$ &
  - &
  69.44$\%$ &
  - &
  - \\\specialrule{0em}{2pt}{2pt} \hline \specialrule{0em}{2pt}{2pt}
\multicolumn{3}{c}{RandLANet} &
  61.66$\%$ &
  74.32$\%$ &
  76.28$\%$ &
  67.68$\%$ &
  87.49$\%$ \\\specialrule{0em}{2pt}{2pt} \hline \specialrule{0em}{2pt}{2pt}
\multicolumn{3}{c}{BAAF-Net} &
  64.66$\%$ &
 77.29$\%$ &
  72.35$\%$ &
  71.86$\%$ &
  \textbf{93.71$\%$} \\ \specialrule{0em}{2pt}{2pt} \hline \specialrule{0em}{2pt}{2pt}
\multicolumn{3}{c}{\textbf{STVP-SD}} &
  \textbf{71.82$\%$} &
    - &
  \textbf{84.30$\%$} &
  \textbf{88.45$\%$} &
  72.64$\%$ \\ \specialrule{0em}{2pt}{2pt} \hline \specialrule{0em}{2pt}{2pt}
\multicolumn{3}{c}{\textbf{STVP}} &
  \textbf{82.09$\%$} &
  \textbf{88.10$\%$} &
  \textbf{90.17$\%$} &
  \textbf{83.72$\%$} &
  \textbf{96.80$\%$} \\ \specialrule{0em}{2pt}{2pt} \hline \specialrule{0em}{2pt}{2pt}
\multicolumn{3}{c}{RS++} &
  58.19$\%$ &
  69.86$\%$ &
  74.06$\%$ &
  69.39$\%$ &
  80.84$\%$ \\ \specialrule{0em}{2pt}{2pt} \hline \specialrule{0em}{2pt}{2pt}
\multicolumn{3}{c}{IDIS++} &
  59.40$\%$ &
  70.90$\%$ &
  79.12$\%$ &
  78.12$\%$ &
  79.53$\%$ \\ \specialrule{0em}{2pt}{2pt} \hline \specialrule{0em}{2pt}{2pt}
\multicolumn{3}{c}{LFA++} &
  61.38$\%$ &
  77.50$\%$ &
  77.32$\%$ &
  81.02$\%$ &
  61.05$\%$\\ \specialrule{0em}{2pt}{2pt} \hline \specialrule{0em}{2pt}{2pt}
\multicolumn{3}{c}{TSD--} &
  67.40$\%$ &
  79.97$\%$ &
  80.95$\%$ &
  81.70$\%$ &
  71.22$\%$\\ \specialrule{0em}{2pt}{2pt} \hline \specialrule{0em}{2pt}{2pt}
\end{tabular}
}
\end{table}

\textbf{The results of saliency detection:}
We compare the saliency prediction performance of STVP-SD with PointNet++, BAFF, and RandLANet.

The experimental results are shown in Table \ref{tb_result}, where STVP-SD achieves performance comparable to or better than state-of-the-art methods in saliency detection. This is because STVP-SD fully utilizes temporal information, spatial coordinates, and color information across point cloud video frames. Among the four performance metrics, STVP-SD performs worse than the other three networks only in recall, due to the more stringent conditions for determining the visible region.

\begin{figure}[htb!]
    \centering
	  \subfloat[]{
      \includegraphics[width=0.25\linewidth]{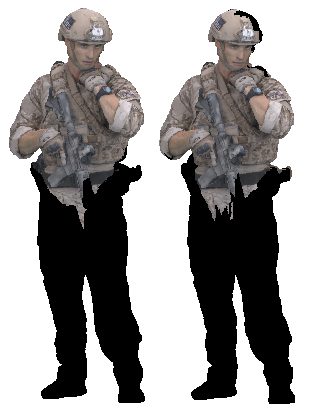} }
      \hspace{0.5mm}
	  \subfloat[]{
        \includegraphics[width=0.25\linewidth]{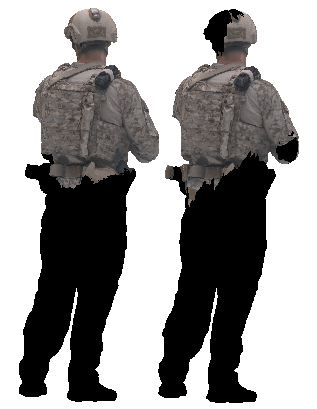}} 
        \hspace{0.5mm}
	  \subfloat[]{
        \includegraphics[width=0.25\linewidth]{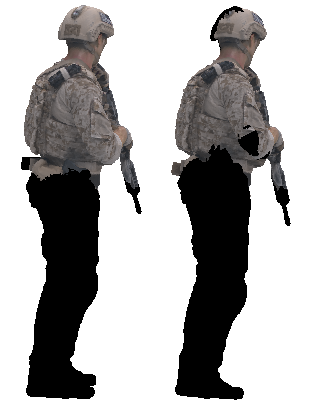}}

	  \subfloat[]{
        \includegraphics[width=0.25\linewidth]{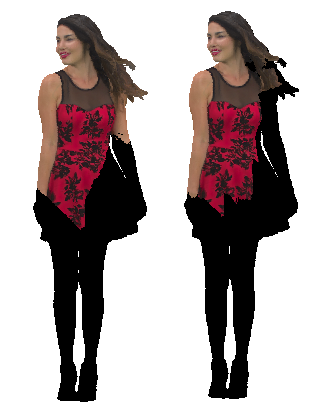}} 
        \hspace{1mm}
	  \subfloat[]{
        \includegraphics[width=0.25\linewidth]{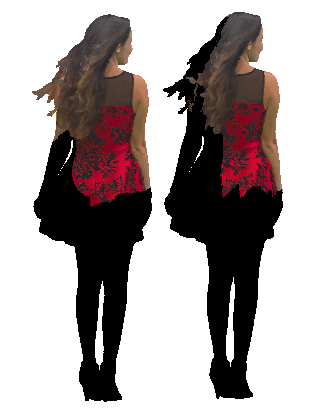}}
        \hspace{1mm}  
         \subfloat[]{
        \includegraphics[width=0.25\linewidth]{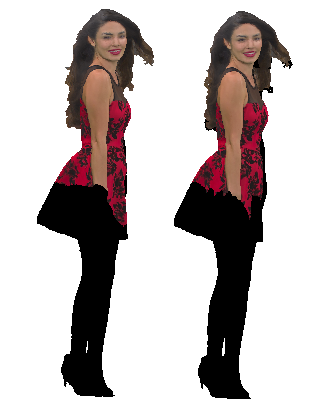}}         
	  \caption{Comparison of predicted results and ground truth in points level for $2$nd frame and $307$th frame. The left, back and side columns are front, back, and side view, respectively. In each subplot, the left side is the prediction result and the right side is the ground truth.}
	  \label{fig_FoV_point_level} 
\end{figure}

\textbf{The results of viewport prediction:}
Table \ref{tb_result} presents the experimental results, showing that STVP achieves high viewport prediction performance, with a point-level MIoU of 82.09\%. The accuracy at the block level is even higher, with an MIoU of 88.10\%.

To visualize the prediction results, we extract one frame from each of two video sequences to display the final viewport prediction. The prediction results are presented in two forms: point-level prediction (Fig. \ref{fig_FoV_point_level}) and block-level prediction (Fig. \ref{fig_FoV_tile_level}). From the point-level prediction results, we observe that although the STVP predictions slightly differ at the edges of the viewport compared to the ground truth, their viewport ranges largely overlap. However, their viewport edges do not exhibit this defect, which is effectively addressed by extending the predictions to the block level. From the block-level prediction results, we observe that the ground truth is almost identical to the predicted result.

\begin{figure}[htb!]
    \centering
      \includegraphics[width=0.9\linewidth]{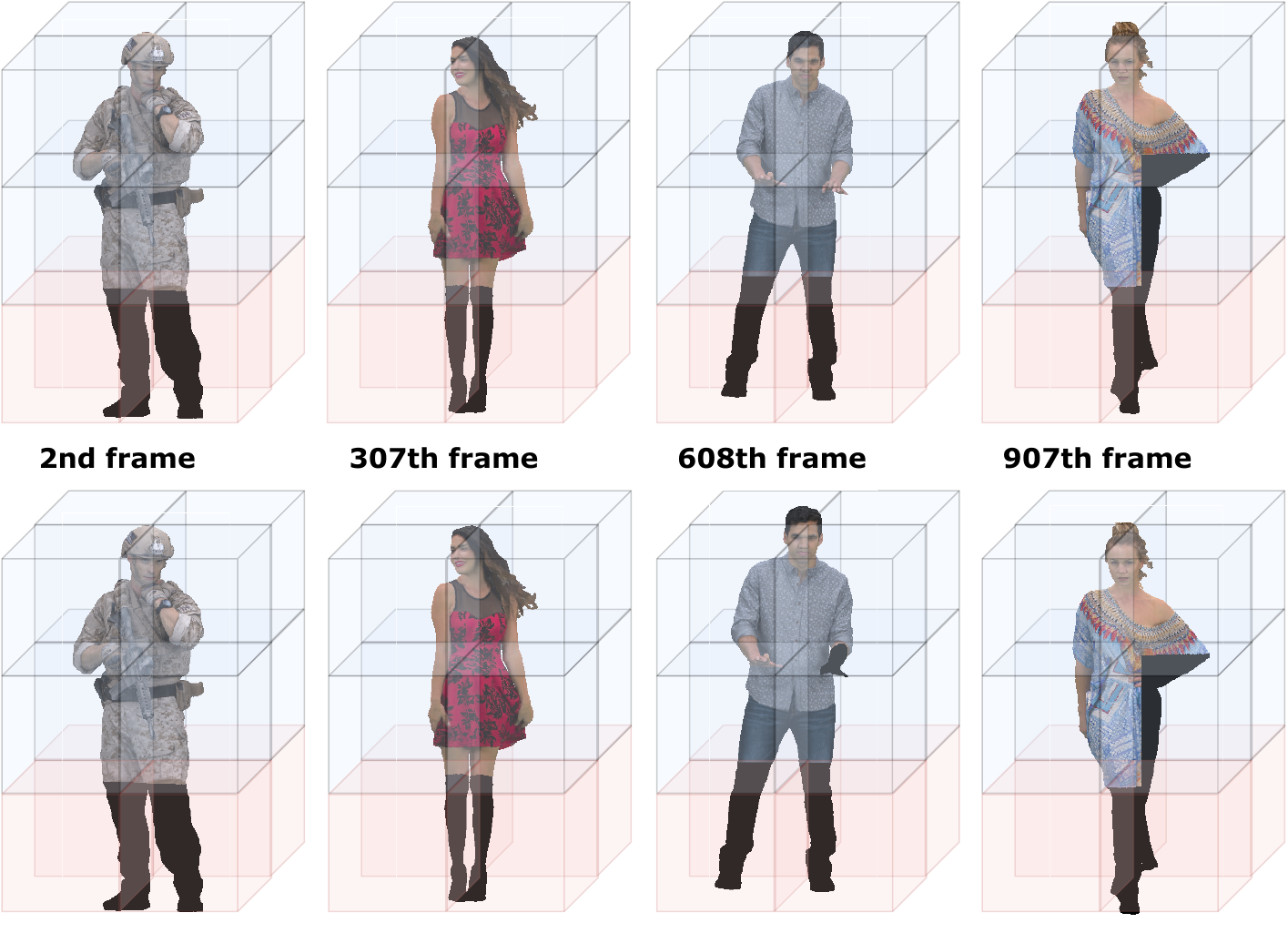} 
	  \caption{Comparison of predicted results and ground truth of four sequences in block-level. The top chart is the prediction chart, and the bottom chart is the corresponding ground truth. }
	  \label{fig_FoV_tile_level} 
\end{figure}

\textbf{The result of ablation:}
We also compare the performance of all ablation networks and present the results in Table \ref{tb_result}.
The result indicate that
substituting the URS unit with either RS or IDIS leads to a reduction of approximately 23\% in the MIoU score of prediction. This shows that URS plays a crucial role in preserving the information of the original point cloud video. Replacing or removing it significantly impacts the viewport prediction.
The LDC replacement also has a large impact on prediction performance.
This is because both spatial color and geometric prominence attract the user's attention. LDC increases color discrepancy coding compared to LFA, which effectively captures spatially salient color regions and significantly improves spatial saliency prediction performance.

We also visualize the prediction results of the ablation experiments in Fig. \ref{fig_ablation}. Since RS does not guarantee that the sampling results of adjacent video frames match, RS++ leads to disconnected frames within the viewport, as shown in Fig.~\ref{fig_ablation1} and Fig. \ref{fig_ablation11}. Due to the low retention of local spatial information in IDIS, IDIS++ also results in the same situation, as shown in Fig.~\ref{fig_ablation2} and Fig. \ref{fig_ablation22}.  Meanwhile, since LFA ignores spatially salient information, from Fig.~\ref{fig_ablation3} and Fig.~\ref{fig_ablation33} we can see that the results of LFA++ have a salient part of the region less than the ground truth, e.g., hair in ``Redandblack'' and gun in ``Soldier''. The temporal detection sub-module is designed to capture the dynamic saliency region between frames. Therefore, as shown in Fig. \ref{fig_ablation4} and Fig. \ref{fig_ablation44}, the result of TSD-- lacks some dynamic salient regions, such as hair in ``Redandblack'' and neck in ``Soldier'', compared to the ground truth. 

\begin{figure}[H]
    \centering
	  \subfloat[ RS++ ]{
      \includegraphics[width=0.14\linewidth]{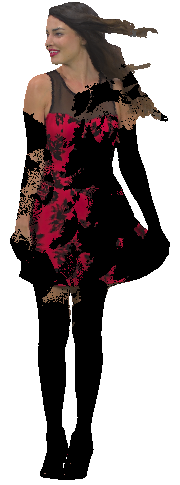}\label{fig_ablation1}  } 
	  \subfloat[IDIS++]{
        \includegraphics[width=0.14\linewidth]{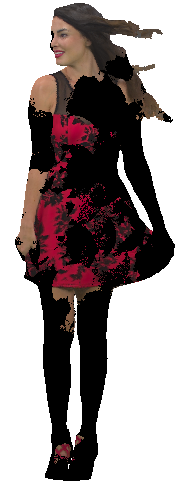}\label{fig_ablation2} } 
         \subfloat[LFA++]{
        \includegraphics[width=0.14\linewidth]{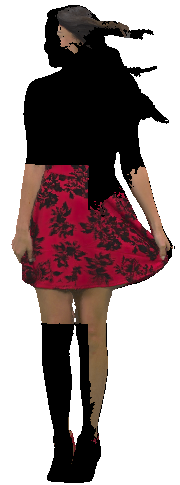}\label{fig_ablation3} } 
         \subfloat[TSD--]{
      \includegraphics[width=0.14\linewidth]{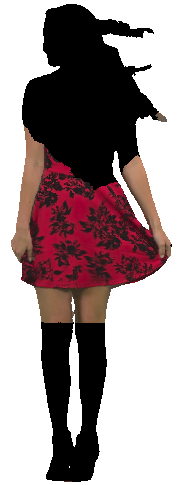}\label{fig_ablation4}}
        \subfloat[STVP]{
      \includegraphics[width=0.14\linewidth]{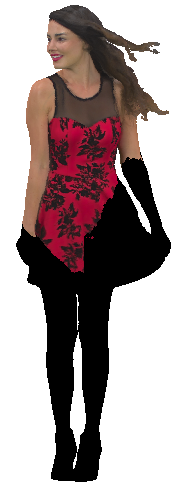}\label{fig_ablation5}} 
        \subfloat[GT]{
      \includegraphics[width=0.14\linewidth]{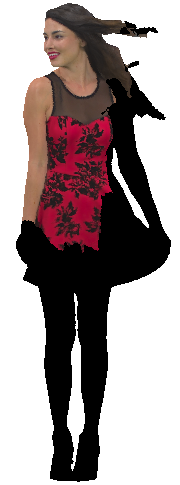}\label{fig_ablation6}}

        	  \subfloat[ RS++ ]{
      \includegraphics[width=0.14\linewidth]{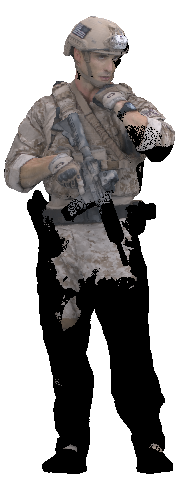}\label{fig_ablation11}  } 
	  \subfloat[IDIS++]{
        \includegraphics[width=0.14\linewidth]{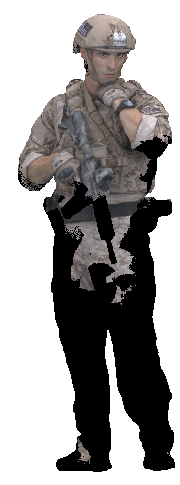}\label{fig_ablation22} } 
         \subfloat[LFA++]{
        \includegraphics[width=0.14\linewidth]{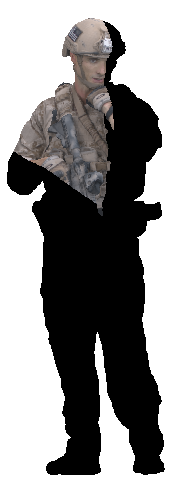}\label{fig_ablation33} } 
         \subfloat[TSD--]{
      \includegraphics[width=0.14\linewidth]{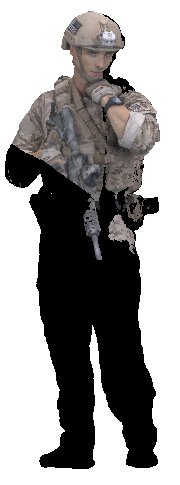}\label{fig_ablation44}}
        \subfloat[STVP]{
      \includegraphics[width=0.14\linewidth]{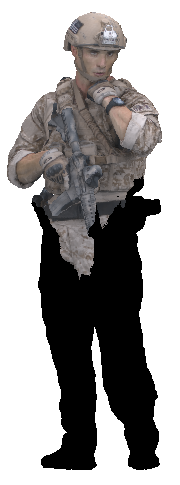}\label{fig_ablation55}} 
        \subfloat[GT]{
      \includegraphics[width=0.14\linewidth]{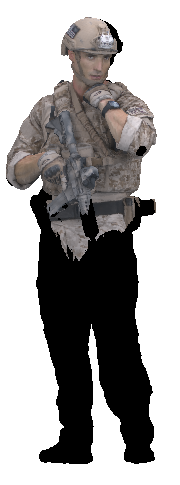}\label{fig_ablation66}} 
        
	  \caption{The predicted results for $311$th  and $11$th frames are compared with the ground truth at the point level and are shown on the images in the first and second rows, respectively. The first four columns of each row indicate the results of each frame in the ablation experiment (1), (2), (3), and (4), the fifth image column denotes the predicted image of our complete network STVP, and the sixth column shows the ground truth. }
	  \label{fig_ablation} 
\end{figure}

\section{Conclusion}
\label{sec:summary}

This paper proposes a high-precision viewport prediction scheme for point cloud videos, leveraging viewport trajectory and video saliency information.
In particular, we propose a novel and efficient sampling
method, along with a new saliency detection method that incorporates temporal and spatial information to capture dynamic, static geometric, and color salient regions.
Our proposal effectively fuses the saliency and trajectory information to achieve more
accurate viewport prediction. We have conducted extensive simulations
to evaluate the efficacy of our proposed viewport prediction
methods on state-of-the-art point cloud video sequences. The
experimental results demonstrate the superiority of our proposal
over existing schemes.

\ifCLASSOPTIONcaptionsoff
  \newpage
\fi


\section*{ACKNOWLEDGMENT}
This work is supported in part by grants from the National Natural Science Foundation of China (52077049), the Anhui Provincial Natural Science Foundation (2008085UD04). 

\bibliography{final_version}
\bibliographystyle{IEEEtran}

\end{document}